\RequirePackage{fix-cm}
\documentclass{article} % For LaTeX2e
\usepackage{iclr2027_conference,times}
\usepackage{microtype}

\usepackage{amsmath,amsfonts,bm}

\def\eqref#1{equation~\ref{#1}}
\def\1{\bm{1}}

\DeclareMathAlphabet{\mathsfit}{\encodingdefault}{\sfdefault}{m}{sl}
\SetMathAlphabet{\mathsfit}{bold}{\encodingdefault}{\sfdefault}{bx}{n}

\usepackage{booktabs}
\usepackage{multirow}
\usepackage{longtable}
\usepackage{amssymb}
\usepackage{amsthm}
\newtheorem{proposition}{Proposition}
\newtheorem{corollary}{Corollary}
\usepackage{algorithm}
\usepackage{algorithmic}
\usepackage{tikz}
\usetikzlibrary{arrows.meta,positioning,fit,backgrounds}
\usepackage[hidelinks]{hyperref}
\usepackage{xurl}
\usepackage{wrapfig} % for wrapfigure
\usepackage{needspace}
\usepackage{fvextra}
\title{MetaDecomp: Adaptive Task Decomposition \\ via Meta-Learned Procedural Memory}

\author{
  Ziyi Wang\textmd{\textsuperscript{1}}\footnotemark[1]\textmd{,}
  ~Yixiao He\textmd{\textsuperscript{1}}{\footnotemark[1]}\textmd{,} 
  ~Yichen Wu\textmd{\textsuperscript{2}}\textmd{,}
  ~Ruite Lu\textmd{\textsuperscript{1}}\textmd{,}
  ~Jiayu Wang\textmd{\textsuperscript{1}}\textmd{,}
  ~Ying Wei\textmd{\textsuperscript{{1}}}\footnotemark[2] \\
  \textsuperscript{1}Zhejiang University, \textsuperscript{2}Harvard University
}

\newif\ifisfinalcopy
\isfinalcopytrue
\ifisfinalcopy
    \iclrfinalcopy
\fi

\begin{document}

\maketitle
\lhead{Preprint}
\vspace{-8pt}

\ifisfinalcopy
    \footnotetext[1]{Equal contribution.}
    \footnotetext[2]{Corresponding author.}
\fi

\begin{abstract}
% Reusing decomposition patterns from past executions can help language agents solve long-horizon tasks, but requires identifying which procedures apply and which steps remain necessary as execution progresses. We introduce MetaDecomp, an in-context meta-learning framework for adaptive task decomposition. Across training tasks, an outer loop jointly learns a state space describing task progress and transition directed acyclic graphs (DAGs) encoding reusable procedures between states. Together, the retained state nodes and transition DAGs form a procedural memory. As the state space evolves, the learner revisits past executions to reassess state assignments and reorganize the procedures they support. Within each task, an inner loop uses this memory to refine an initial decomposition and revises it using environmental feedback. At test time, the learned memory remains fixed while the decomposition adapts to each task. Evaluated over three seeds on ScienceWorld and ALFWorld, MetaDecomp achieves the highest mean performance among the compared methods in four of five benchmark--model settings, using fewer inference tokens than ReAct and AutoRefine in all five settings. An ablation study further examines the roles of the inner and outer loops. Code will be released.

Reusing past experience in long-horizon tasks requires more than storing successful procedures: Language agents must distinguish which procedures apply and what work remains as execution progresses. We introduce MetaDecomp, an in-context meta-learning framework for adaptive task decomposition that jointly learns a task-relative state space and reusable procedures. Its outer loop builds procedural memory from successful training executions, representing task progress as abstract states and procedures between states as directed acyclic graphs of operations and dependencies. As new progress distinctions are learned, MetaDecomp revisits historical execution evidence to reassess state assignments and reorganize the procedures they support. With memory fixed at test time, the inner loop refines an initial decomposition and revises it from environmental feedback, combining applicable procedure fragments with task-specific steps missing from memory. Evaluated on ALFWorld and ScienceWorld over three seeds, MetaDecomp achieves the highest mean performance among the compared methods in four of five settings, for example, an 86.3\% success rate on ALFWorld with Qwen3-4B. Fixing the initial schema or omitting historical reprojection reduces performance in component ablations. MetaDecomp also uses fewer inference tokens than ReAct and AutoRefine in all five settings. Cross-benchmark analysis identifies a low-performance area where partial coverage degrades success relative to hidden memory; high coverage improves success, while low coverage changes little. Code will be released upon acceptance.
\end{abstract}

\section{Introduction}

\begin{wrapfigure}{r}{0.48\textwidth}
\centering
\vspace{-3em}
\includegraphics[width=\linewidth]{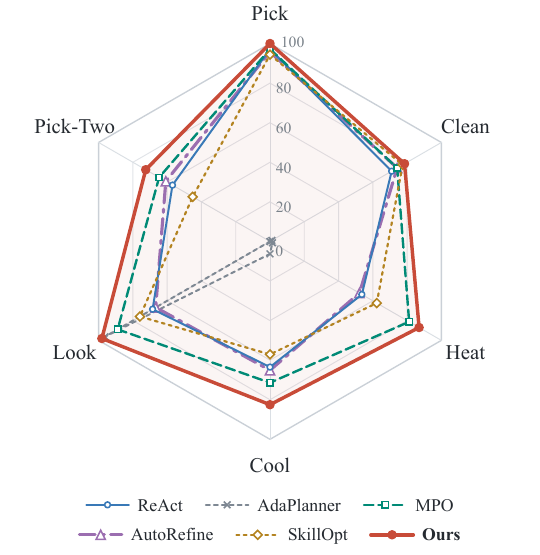}
\caption{Mean success rates (\%) over three seeds on ALFWorld with Qwen3-4B (Table~\ref{tab:main_results_alfworld}).}
\label{fig:alfworld-radar}
\vspace{-2em}
\end{wrapfigure}

Task decomposition helps language agents organize long-horizon tasks into intermediate subgoals and their dependencies~\citep{huang2022zeroshotplanner,khot2023decomposed}. Across related task instances, patterns such as acquiring an object, preparing it for use, and placing it at a destination recur even when concrete entities differ. Learning these patterns from past executions can provide reusable guidance for new tasks. Effective reuse, however, requires deciding which procedures apply and which parts remain necessary given the task requirements and the progress already established.

Prior work learns reusable knowledge in the form of transferable insights, executable skills, and workflows, as in ExpeL, Voyager, and Agent Workflow Memory~\citep{zhaoEtAl2024ExpeL,wangEtAl2024Voyager,wangEtAl2025AgentWorkflowMemory}. Recent systems organize procedures through hierarchical workflows and skill graphs~\citep{fanEtAl2025MetaFlowLLM,xuEtAl2026HyperSkill}, while HiSkill associates skills with applicability conditions and expected state changes~\citep{haoEtAl2026HiSkill}. These advances motivate a complementary question for decomposition reuse: which aspects of task progress should a shared representation retain to organize procedures from different executions?

Consider cleaning-and-placement tasks in ALFWorld~\citep{shridharEtAl2021ALFWorld}. A mug and a plate can share a procedure for acquisition, cleaning, and placement. An agent already holding a clean object, however, should proceed to placement, whereas one holding a dirty object still needs to clean it. A representation that records possession alone merges histories requiring different continuations. Retaining cleanliness can distinguish these continuations while allowing procedures to transfer across concrete objects. Learning abstractions for planning connects reusable actions to the conditions needed to compose them~\citep{konidaris2018skills,bonet2020representations}. For decomposition reuse, this connection motivates learning the representation of progress together with the procedures it organizes.

\begin{wrapfigure}{R}{0.43\textwidth}
\centering
\vspace{-1.5em}
\includegraphics[width=\linewidth]{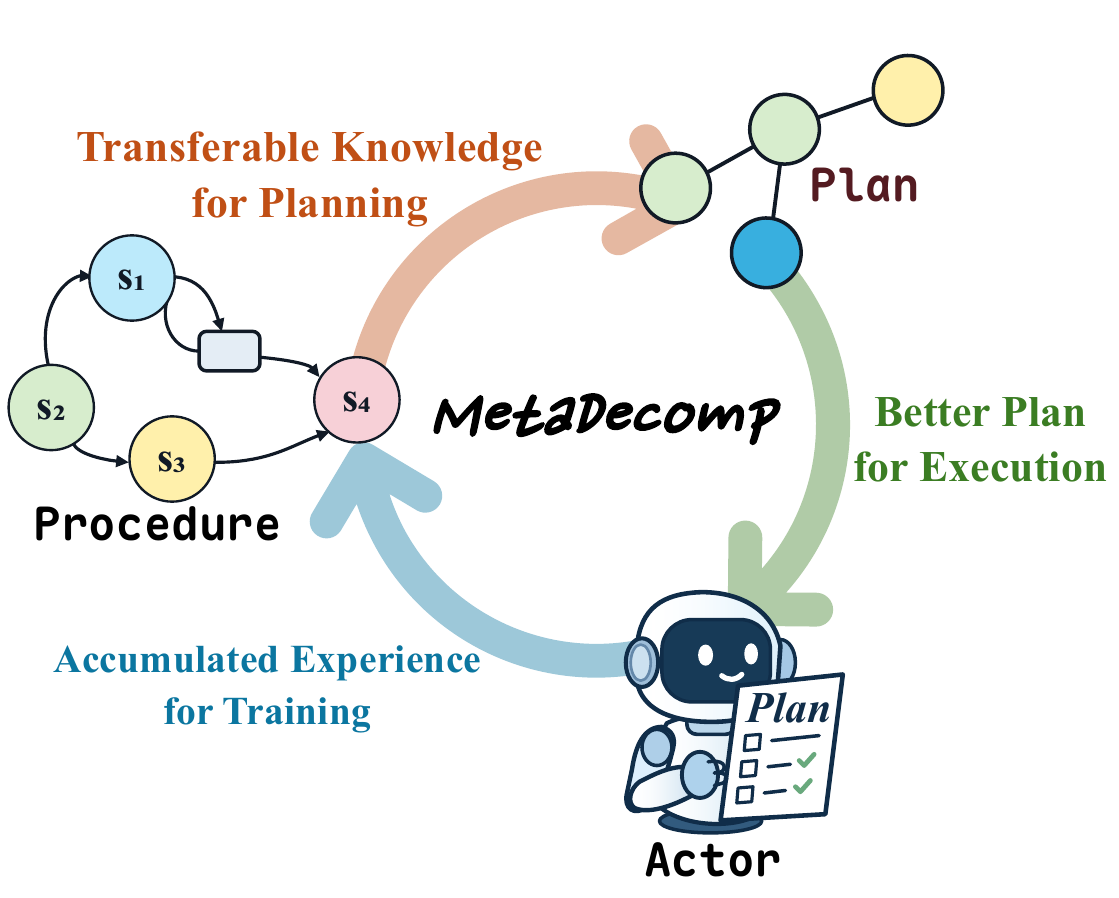}
\caption{Learning to decompose from experience. Shared procedural knowledge guides task-specific planning and execution. Successful training executions update that knowledge for later tasks.}
\label{fig:overview}
\vspace{-0.3em}
\end{wrapfigure}

\leavevmode We introduce MetaDecomp, an in-context meta-learning framework that learns a shared procedural prior across related tasks and uses it to adapt task-specific decompositions from environmental feedback. This prior jointly represents task progress and reusable procedures through a learned state space and transition DAGs. States capture abstract progress conditions, separating what has been established from the specific entities involved. Each transition DAG encodes a reusable procedure between two states as a directed acyclic graph of operations and dependencies. Together, the retained state nodes and transition DAGs form a procedural memory, represented as a directed state multigraph whose edges carry transition DAGs. Multiple procedures may connect the same state pair.

Learning proceeds on two timescales, separating knowledge accumulated across related tasks from adaptation within each task~\citep{santoro2016memory,hospedales2022metalearning}. The outer loop jointly learns the state space and transition DAGs from successful training executions. As new progress distinctions are introduced, it revisits the original execution evidence to reassess historical state assignments and the procedures supported at each transition. Changes to the state space therefore reorganize the procedures in this memory.

The inner loop uses this shared knowledge to guide the initial decomposition and revises it as execution unfolds. Building on feedback-driven planning~\citep{sun2023adaplanner,prasadEtAl2024ADaPT}, it interprets observed progress in the learned state space and recalls procedures that may help resolve unfinished work. The planner can combine parts of stored procedures and add work absent from the procedural memory. Successful training executions return to the outer loop, completing the learning cycle in Figure~\ref{fig:overview}. At test time, the learned state space and transition DAGs remain fixed while the decomposition adapts to each task.

Our contributions are as follows:
\begin{itemize}
    \item \textbf{MetaDecomp learns how to decompose across tasks rather than starting from scratch on each one (Section~\ref{sec:method}).} In the outer loop, it distills a shared procedural prior from executions on related tasks. In the inner loop, it uses that prior to build and revise a decomposition for the current task based on environmental feedback.
    \item \textbf{Reusable procedures are anchored to explicit task progress (Sections~\ref{sec:outer}--\ref{sec:adaptation} and Table~\ref{tab:component-ablation}).} We learn progress abstractions and transition DAGs together, so each procedure links what has already been accomplished to how the remaining work can be done. As the abstractions are refined, we reinterpret past executions so that these links stay consistent.
    \item \textbf{MetaDecomp is both stronger and cheaper in most settings, and we identify when transfer stops helping (Section~\ref{sec:experiments}).} It achieves the highest mean performance in four of the five ALFWorld/ScienceWorld settings while using fewer inference tokens than ReAct and AutoRefine in all five. Across eight additional benchmarks, our transfer analysis shows that reuse helps when procedures are highly applicable, but leads to negative transfer when they only partially apply.
\end{itemize}

\section{Related Work}

\paragraph{Meta-Learning.}
Meta-learning uses experience across tasks to support adaptation~\citep{hospedales2022metalearning}. Classical approaches learn parameter initializations or memory-based adaptation mechanisms~\citep{finn2017maml,santoro2016memory}. Language-agent methods learn meta-planners from execution feedback~\citep{xiong2025mpo}, train memory-management policies~\citep{liangEtAl2026MCMA}, or evolve memory architectures~\citep{zhangEtAl2025MemEvolve}. MetaDecomp learns a shared procedural prior of progress abstractions and reusable procedures, adapting each task's decomposition with this prior and environmental feedback.

\paragraph{Agent Memory.}
Agents accumulate executable skills~\citep{wangEtAl2024Voyager}, workflow subroutines in Agent Workflow Memory (AWM)~\citep{wangEtAl2025AgentWorkflowMemory}, and playbooks or skill documents~\citep{zhangEtAl2026ACE,yang2026skillopt}. A-MEM revises linked memories~\citep{xuEtAl2025AMEM}; MACLA learns hierarchical procedures with preconditions and postconditions~\citep{forouzandehEtAl2025MACLA}. MemP and ReMe reuse procedural experience~\citep{fangEtAl2026MemP,caoEtAl2025ReMe}; SkillDAG and HyperSkill organize skills as graphs or hypergraphs~\citep{baiEtAl2026SkillDAG,xuEtAl2026HyperSkill}. Beyond AWM's workflow induction, MetaDecomp reprojects original executions when its progress schema changes and reassesses the procedures supported at each state pair.

\paragraph{Planning and Decomposition.}
Planning-model learning studies abstractions for supplied skills~\citep{konidaris2018skills} and first-order symbolic models~\citep{bonet2020representations}; MetaDecomp jointly learns progress abstractions and procedures. TheoryCoder-2 learns abstractions and an executable dynamics model for hierarchical planning~\citep{ahmed2026theorycoder2}. MetaDecomp instead uses an LLM to ground and compose candidate procedures and supply missing steps. Its graph is not a recursive hierarchical task-network solver~\citep{erol1994htn}. AdaPlanner and ADaPT revise or decompose plans using execution feedback~\citep{sun2023adaplanner,prasadEtAl2024ADaPT}.

\section{MetaDecomp}
\label{sec:method}

\begin{figure}[t]
    \centering
    \includegraphics[width=\linewidth]{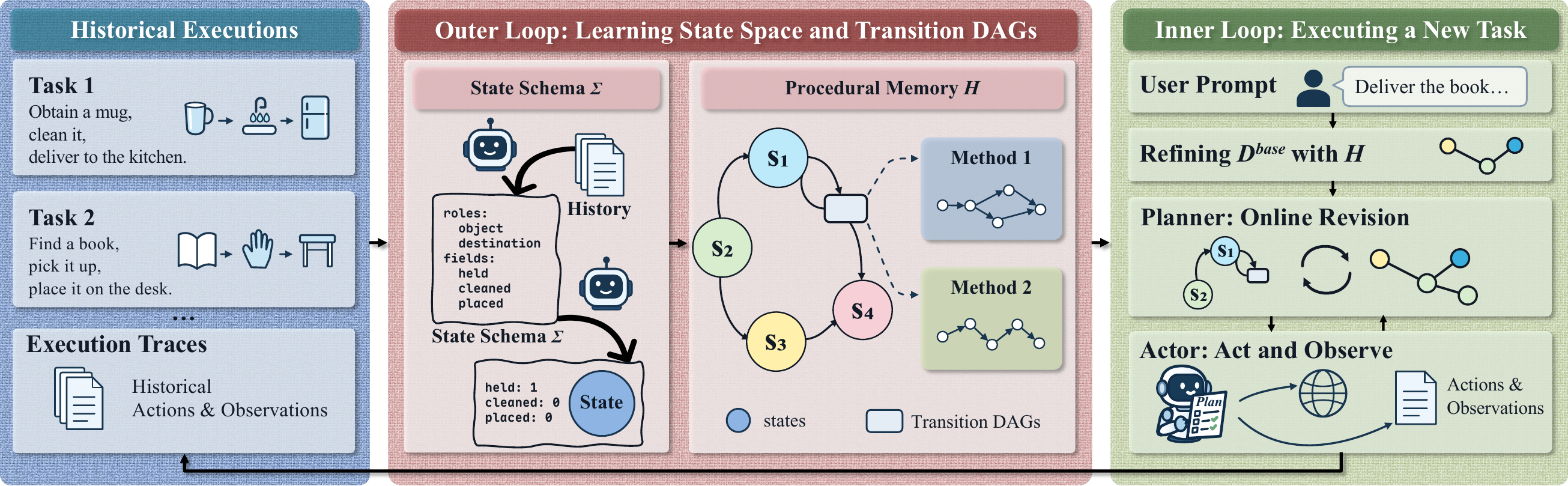}
    \caption{MetaDecomp jointly learns a state space specified by schema $\Sigma$ and transition DAGs in that space. Retained state nodes and transition DAGs form the procedural memory $\mathcal H$. The inner loop refines a cached decomposition $D^{\mathrm{base}}$ into $D_0$ and revises it using actor feedback. Successful training executions update the shared knowledge between episodes; evaluation keeps it fixed. The illustrative binary fields denote established (1) or unestablished (0) progress.}
    \label{fig:pipeline}
\end{figure}

MetaDecomp is an in-context meta-learning framework for task decomposition with two coupled loops. The outer loop jointly learns a state space describing task progress and transition DAGs in that space from executions across related tasks. The retained state nodes and transition DAGs form a procedural memory, which the inner loop uses to construct and adapt a decomposition for the current task. Successful training executions supply evidence for the next outer update. Figure~\ref{fig:pipeline} illustrates this cycle, and Algorithm~\ref{alg:metadecomp} gives the complete procedure.

\subsection{Preliminaries}
\label{sec:representation}

We consider interactive tasks organized into learning spaces. Each learning space covers a supplied task type or a predefined group of related types within one benchmark and learns its own state space and transition DAGs. For task $\tau$, an actor receives observations and executes environment actions, producing a history $h_t=(o_0,a_1,o_1,\ldots,a_t,o_t)$. Its guidance is an ordered decomposition $D_t=(d_{t,1},\ldots,d_{t,n_t})$ of task-level instructions. The state space and transition DAGs transfer across instances within the learning space, while $D_t$ is constructed and adapted for the current task.

Consider cleaning and placing an object. A mug and a plate can share a procedure even though their concrete identities differ. Whether the required object has already been cleaned, however, changes the remaining work. MetaDecomp therefore separates task-relative roles, such as the required object, from progress conditions, such as possession and cleanliness. A learned schema $\Sigma$ specifies the roles and a set of fields $\mathcal F$, each with a finite value set $V_f$. It defines the state space $\mathcal X_\Sigma$, on which the procedural memory $\mathcal H$ is constructed,
\begin{equation}
 \mathcal X_\Sigma=\prod_{f\in\mathcal F}V_f,\qquad
 \mathcal H=(\mathcal S,\mathcal T),\qquad
 \mathcal S\subseteq\mathcal X_\Sigma.
 \label{eq:schema}
\end{equation}
A node $s\in\mathcal S$ is a complete field assignment retained from experience, and $\mathcal T$ contains transitions between these nodes. Role bindings $\beta$ associate abstract referents with task entities and do not enter state identity. Two tasks can therefore share a state recording that the required object is held and clean while binding that object to different entities. Field values describe conditions supported by observations, including explicit values for unknown or unestablished progress.

Each transition $e\in\mathcal T$ connects two distinct states through a transition DAG $M_e$ representing a reusable procedure, which we call a method,
\begin{equation}
 e=(s_e^-,s_e^+,M_e),\qquad
 M_e=(A_e,E_e,x_e),\qquad E_e\subseteq A_e\times A_e.
 \label{eq:procedural-memory}
\end{equation}
The nodes $A_e$ are parameterized operations, $x_e(a)$ gives the instruction for operation $a$, and the acyclic edges $E_e$ express inferred prerequisite dependencies. Instructions retain supported applicability, argument-selection, and stopping conditions. Multiple methods can realize the same state change, and the state graph may contain cycles. The state nodes and transition DAGs together form the procedural memory, a directed state multigraph whose transitions carry these DAGs. Procedures from different task instances can thus be indexed by the same progress conditions, independently of concrete entity bindings.

Let $\Sigma_b$ define the state space and $\mathcal H_b$ be the procedural memory after outer update $b$. An evidence ledger $\mathcal P_b$ retains executions and their state and method assignments. For language-model modules $\phi$, the two-loop process is
\begin{equation}
 \begin{aligned}
 \xi_\tau&=\operatorname{Inner}_\phi(\tau;\Sigma_b,\mathcal H_b),\\
 (\Sigma_{b+1},\mathcal H_{b+1},\mathcal P_{b+1})
 &=\operatorname{Outer}_\phi(\Sigma_b,\mathcal H_b,\mathcal P_b,\mathcal B_b),
 \end{aligned}
 \label{eq:two-levels}
\end{equation}
where $\xi_\tau=(\tau,h_{T_\tau})$ is an executed episode and $\mathcal B_b$ is a batch of successful training episodes. The outer loop updates the shared schema and transition DAGs in external memory while language-model weights remain fixed. The inner loop constructs and revises $D_t$ for one task with $(\Sigma_b,\mathcal H_b)$ fixed. The shared procedural prior thus supports task adaptation through changes to the decomposition supplied in context.

\subsection{Outer loop: jointly learning the state space and transition DAGs}
\label{sec:outer}

The outer loop jointly learns the state space and transition DAGs from batches of successful training executions. Task descriptions specify requirements, and action--observation traces supply evidence of attained conditions and executed operations. The learner first proposes progress distinctions needed to determine what remains to be done, which methods apply, or whether the task goal is satisfied. The schema can acquire new roles, fields, or values while retaining the meanings of existing entries. These additions determine how execution experience is organized in the procedural memory.

Under the updated schema, the learner extracts local realizations of procedures from each new execution. A boundary identifies the entry observation or the observation after an executed action. Each realization $\ell$ contains a source boundary $\omega_\ell^-$, a later target boundary $\omega_\ell^+$, and evidence-supported operations proposed to connect their states. When the schema changes, the learner also returns to the original traces and reprojects retained historical boundaries. It reconsiders only added or expanded fields and new role bindings, preserving the other assignments and execution evidence. Writing $q_{b+1}$ for the resulting boundary projection, new and reassessed realizations are grouped by their complete endpoint assignments,
\begin{equation}
 \mathcal E_{uv}=\{\ell\in\mathcal L_b:
 q_{b+1}(\omega_\ell^-)=u,\ q_{b+1}(\omega_\ell^+)=v\},\qquad u\ne v,
 \label{eq:experience-groups}
\end{equation}
where $\mathcal L_b$ contains the realizations under review at update $b$. Equality depends on progress conditions, so different entity bindings can contribute evidence to the same state change.

Changing these groups also changes the support of existing methods. A method retains its association with one state pair only when all its supporting realizations agree on that pair after reprojection. If those realizations map to different state pairs, the learner reassesses the procedure separately for each resulting pair of distinct retained states. Appendix~\ref{app:schema-reassignment-case} illustrates how adding cleanliness separates the evidence supporting cleaning-and-placement procedures. The stored links to original evidence make this reassignment possible. This couples learning the state space with learning transition DAGs in that space. A new progress distinction changes which experiences can support a common method and how that method connects states.

To update the transition DAGs, the learner consolidates realizations that support the same parameterized procedure for each state pair under review. It can refine instructions, add evidenced operations, or revise dependencies. Procedures with different operations, applicability, argument-selection or stopping conditions, or inferred dependencies remain separate alternatives. New operations require execution support, and proposed DAGs are checked for valid references and acyclicity. At reuse, the planner binds method roles to task entities and checks applicability using the task and any available execution observations. The updated state space and transition DAGs are available to later training tasks. Appendix~\ref{app:learning} specifies evidence extraction, historical reassignment, and method updates.

\subsection{Inner loop: task adaptation with procedural memory}
\label{sec:adaptation}

The inner loop interprets task progress in the learned state space and uses the stored transition DAGs as a procedural prior. The planner binds task roles, selects useful operations, and combines them with any work absent from the memory into an ordered decomposition, which guides the actor in choosing concrete environment commands based on the current observations.

In the main experimental configuration, MetaDecomp starts from MPO's cached task-specific decomposition $D^{\mathrm{base}}$. Initial refinement aligns this decomposition with the learned state space and procedures. The planner receives all methods for the task's learning space, without requiring a stored-state match, and constructs
\begin{equation}
 D_0=\operatorname{Plan}_\phi(\tau,D^{\mathrm{base}};\Sigma,\mathcal H).
 \label{eq:initial-plan}
\end{equation}
The planner can replace, insert, or remove instructions to incorporate applicable experience and repair missing or unsuitable work. It identifies each reused operation's prerequisites and places any steps needed to establish them earlier in the plan. When the procedural memory contains no methods, initialization preserves the cached decomposition in dependency order.

MPO is an experimental initializer rather than a required component of the representation or outer learner. Section~\ref{sec:initialization} evaluates task-only, ReWOO, and ADaPT initialization.

When execution feedback indicates stalled progress, the planner reviews the current decomposition. When a schema is available, it first projects the observed execution history into that schema. Online path recall requires a complete projection with exactly the field values of a stored state. The model selects intermediate or completion targets from states reachable from the matched current state. MetaDecomp then recalls a bounded number of shortest simple paths to those targets, retaining all alternative methods on their edges. Paths are ranked by state-transition count. The resulting procedures propose ways to continue, whose applicability is judged using the task and current observations.

Let $c_t$ collect the execution evidence, stall information, any complete current state projection, and the cached baseline when available. Let $\mathcal R_t$ denote the recalled methods. Plan revision and the next action follow
\begin{equation}
 D_t^+=\operatorname{Refine}_\phi(\tau,D_t,c_t;\Sigma,\mathcal R_t),\qquad
 a_{t+1}\sim\pi_\phi(\cdot\mid\tau,h_t,D_t^+).
 \label{eq:adapt-and-act}
\end{equation}
The revision model identifies unresolved work from the observations and can replace an instruction, insert work around an existing step, or retain the plan. The planner may reuse part of a method when prerequisites for the reused operations already hold or will be established by earlier planned work. For example, evidence that the object is held but still requires cleaning can motivate inserting cleaning before placement. Revision from execution evidence remains possible when no methods are recalled or applicable. Dependencies guide the model's construction of the sequence, with the actor continuing to ground its instructions as observations arrive.

During training, successful inner-loop executions enter the next outer update, closing the meta-learning cycle in Equation~\ref{eq:two-levels}. During evaluation, each learning space uses a frozen snapshot of its learned state space and transition DAGs and runs only the inner loop. The agent continues to adapt its decomposition to each test task while keeping $\Sigma$ and $\mathcal H$ fixed.

\subsection{Motivation from state abstraction}
\label{sec:interface-analysis}

State abstraction motivates retaining distinctions that affect procedural choices~\citep{li2006abstraction}. Consider a fixed method dictionary, continuation loss, and distribution of histories and task context. If a refined representation preserves the old state, an optimal selector can ignore its added information, so the minimum expected loss cannot increase. It decreases strictly when a coarse-state optimal choice is suboptimal under the refined information with positive probability. Appendix~\ref{app:theory} formalizes this observation and relates unresolved choice ambiguity to excess loss under a unique history-optimal method and a uniform positive margin.

The comparison isolates the value of additional progress information when the method dictionary is fixed. MetaDecomp also changes state assignments and procedures, while finer states can reduce exact-match recall coverage; Section~\ref{sec:loop-ablation} therefore evaluates schema expansion and historical reprojection within the full learning pipeline.

\setlength{\parskip}{4.5pt plus 0.6pt minus 0.6pt}
\setlength{\intextsep}{8pt plus 2pt minus 2pt}
\section{Experiments}
\label{sec:experiments}

We examine how the state space and transition DAGs learned across tasks support adaptation to new task instances, evaluating task completion and inference cost.

\subsection{Experimental Setup}
\label{sec:experimental-setup}

We evaluate on ALFWorld~\citep{shridharEtAl2021ALFWorld} and ScienceWorld~\citep{wangEtAl2022ScienceWorld}. ALFWorld tests household manipulation through a text interface. We evaluate on 134 unseen tasks across six types and report success rate. ScienceWorld requires agents to carry out interactive scientific experiments. We use a fixed panel of 211 test tasks spanning 30 native task types, grouped into ten categories, and report their mean score. In both benchmarks, Avg. is the mean over all test tasks, weighting categories by their test-task counts. Tables~\ref{tab:main_results_alfworld} and~\ref{tab:main_results} report means and sample standard deviations over three evaluation seeds.

For the main benchmark comparisons, ALFWorld uses Qwen3-4B and Qwen3-8B~\citep{qwen3technicalreport}. ScienceWorld uses Qwen3-8B, Qwen3.5-9B, and Qwen3.5-27B~\citep{qwen3.5}. Across all experiments, MetaDecomp's inner loop uses the reported actor model for initial refinement, state projection, target selection, plan revision, and action generation.

We compare with ReAct~\citep{yao2023react}, AdaPlanner~\citep{sun2023adaplanner}, MPO~\citep{xiong2025mpo}, AutoRefine~\citep{qiu2026autorefinecompilingtrajectoriesvalidated}, and SkillOpt~\citep{yang2026skillopt}. Across all MetaDecomp experiments, GLM-5.3-Flash~\citep{zai2026glm53flash} is used only for outer-loop memory learning. AutoRefine and SkillOpt use it for experience distillation during training. During evaluation, AutoRefine and SkillOpt freeze their learned libraries, while MetaDecomp freezes its state space and transition DAGs. Appendix~\ref{app:implementation} provides training and evaluation protocols, category sizes (Table~\ref{tab:evaluation-counts}), and implementation settings.

In the main experiments, training tasks are sampled randomly, and each learning space uses two successful episodes per regular memory update. Checkpoints are selected by validation performance and frozen for testing. Tables~\ref{tab:training-summary} and~\ref{tab:memory-summary} report training resources, memory sizes, and test-time reuse. Test-time reuse refers to inclusion in accepted initial plans or applied revisions.

\subsection{Main Results}
\label{sec:main-results}
\label{sec:inference-efficiency}

Tables~\ref{tab:main_results_alfworld} and~\ref{tab:main_results} show that MetaDecomp achieves the highest mean performance in four of the five benchmark--model settings.

\providecommand{\mainmeansd}[2]{#1\textsubscript{\tiny$\pm$#2}}
\begin{table}[t]
\centering
\small
\setlength{\tabcolsep}{1.5pt}
\renewcommand{\arraystretch}{1.10}
\begin{tabular*}{\textwidth}{@{\extracolsep{\fill}}l*{7}{c}@{}}
\toprule
Method & Pick & Clean & Heat & Cool & Look & Pick-Two & Avg. \\
\midrule
\multicolumn{8}{c}{Qwen3-8B} \\
\midrule
ReAct & \mainmeansd{\underline{98.6}}{2.4} & \mainmeansd{\textbf{90.3}}{0.0} & \mainmeansd{60.9}{8.7} & \mainmeansd{87.3}{2.7} & \mainmeansd{88.9}{0.0} & \mainmeansd{39.2}{12.2} & \mainmeansd{79.6}{2.6} \\
AdaPlanner & \mainmeansd{0.0}{0.0} & \mainmeansd{3.2}{0.0} & \mainmeansd{0.0}{0.0} & \mainmeansd{0.0}{0.0} & \mainmeansd{85.2}{3.2} & \mainmeansd{0.0}{0.0} & \mainmeansd{12.2}{0.4} \\
MPO & \mainmeansd{\textbf{100.0}}{0.0} & \mainmeansd{\underline{88.2}}{1.9} & \mainmeansd{\underline{84.1}}{6.6} & \mainmeansd{\underline{88.9}}{2.7} & \mainmeansd{\underline{96.3}}{3.2} & \mainmeansd{\underline{84.3}}{9.0} & \mainmeansd{\underline{90.3}}{0.7} \\
AutoRefine$^{\dagger}$ & \mainmeansd{\textbf{100.0}}{0.0} & \mainmeansd{\textbf{90.3}}{0.0} & \mainmeansd{66.7}{6.6} & \mainmeansd{87.3}{7.3} & \mainmeansd{87.0}{3.2} & \mainmeansd{33.3}{6.8} & \mainmeansd{79.9}{2.2} \\
SkillOpt$^{\dagger}$ & \mainmeansd{\textbf{100.0}}{0.0} & \mainmeansd{83.9}{0.0} & \mainmeansd{\textbf{92.8}}{2.5} & \mainmeansd{87.3}{2.7} & \mainmeansd{85.2}{3.2} & \mainmeansd{56.9}{13.6} & \mainmeansd{85.6}{1.7} \\
\midrule
Ours  & \mainmeansd{\textbf{100.0}}{0.0} & \mainmeansd{\textbf{90.3}}{3.2} & \mainmeansd{\textbf{92.8}}{2.5} & \mainmeansd{\textbf{95.2}}{0.0} & \mainmeansd{\textbf{100.0}}{0.0} & \mainmeansd{\textbf{90.2}}{3.4} & \mainmeansd{\textbf{94.5}}{0.4} \\
\midrule
\multicolumn{8}{c}{Qwen3-4B} \\
\midrule
ReAct & \mainmeansd{\underline{97.2}}{2.4} & \mainmeansd{71.0}{6.5} & \mainmeansd{53.6}{6.6} & \mainmeansd{63.5}{2.7} & \mainmeansd{68.5}{6.4} & \mainmeansd{56.9}{3.4} & \mainmeansd{69.4}{1.3} \\
AdaPlanner & \mainmeansd{0.0}{0.0} & \mainmeansd{1.1}{1.9} & \mainmeansd{1.4}{2.5} & \mainmeansd{6.3}{2.7} & \mainmeansd{\underline{96.3}}{3.2} & \mainmeansd{0.0}{0.0} & \mainmeansd{14.4}{0.9} \\
MPO & \mainmeansd{\underline{97.2}}{4.8} & \mainmeansd{74.2}{3.2} & \mainmeansd{\underline{81.2}}{2.5} & \mainmeansd{\underline{71.4}}{4.8} & \mainmeansd{88.9}{0.0} & \mainmeansd{\underline{64.7}}{0.0} & \mainmeansd{\underline{79.9}}{2.7} \\
AutoRefine$^{\dagger}$ & \mainmeansd{95.8}{0.0} & \mainmeansd{74.2}{3.2} & \mainmeansd{52.2}{7.5} & \mainmeansd{65.1}{9.9} & \mainmeansd{66.7}{14.7} & \mainmeansd{60.8}{9.0} & \mainmeansd{70.1}{2.7} \\
SkillOpt$^{\dagger}$ & \mainmeansd{94.4}{6.4} & \mainmeansd{\underline{77.4}}{0.0} & \mainmeansd{62.3}{2.5} & \mainmeansd{57.1}{4.8} & \mainmeansd{75.9}{3.2} & \mainmeansd{45.1}{12.2} & \mainmeansd{70.4}{2.6} \\
\midrule
Ours  & \mainmeansd{\textbf{100.0}}{0.0} & \mainmeansd{\textbf{78.5}}{4.9} & \mainmeansd{\textbf{87.0}}{7.5} & \mainmeansd{\textbf{82.5}}{2.7} & \mainmeansd{\textbf{98.1}}{3.2} & \mainmeansd{\textbf{72.5}}{3.4} & \mainmeansd{\textbf{86.3}}{2.6} \\
\bottomrule
\end{tabular*}

\caption{ALFWorld success rates (\%) on 134 unseen test tasks, reported as the mean with sample standard deviation in subscript across three evaluation runs. Avg. weights task types by their test-task counts within each run. Bold and underlining mark the highest and second-highest distinct displayed means, including ties. $^{\dagger}$GLM-5.3-Flash is used for distillation only during training.}
% \vspace{-0.5em}
\label{tab:main_results_alfworld}
\end{table}

On ALFWorld, MetaDecomp leads with both actor sizes; Figure~\ref{fig:alfworld-radar} shows the Qwen3-4B comparison across task types. Figure~\ref{fig:qualitative-memory} illustrates how learned progress distinctions guide the choice of remaining procedures (Appendix~\ref{app:qualitative-memory}).

\providecommand{\mainmeansd}[2]{#1\textsubscript{\tiny$\pm$#2}}
\begin{table}[t]
\centering
\fontsize{8.5}{10}\selectfont
\setlength{\tabcolsep}{0.5pt}
\renewcommand{\arraystretch}{1.10}
\begin{tabular*}{\textwidth}{@{\extracolsep{\fill}}l*{11}{c}@{}}
\toprule
Method & Matter & Meas. & Elec. & Class. & Plants & Chem. & Life-1 & Life-2 & Force & Mende. & Avg. \\
\midrule
\multicolumn{12}{c}{Qwen3-8B~\citep{qwen3technicalreport}} \\
\midrule
ReAct & \mainmeansd{24.8}{7.7} & \mainmeansd{66.3}{3.8} & \mainmeansd{57.6}{1.6} & \mainmeansd{79.5}{3.4} & \mainmeansd{20.4}{6.2} & \mainmeansd{36.5}{2.3} & \mainmeansd{65.6}{8.1} & \mainmeansd{27.2}{11.2} & \mainmeansd{37.5}{6.0} & \mainmeansd{\textbf{40.3}}{13.2} & \mainmeansd{48.3}{1.3} \\
AdaPlanner & \mainmeansd{2.4}{1.2} & \mainmeansd{17.1}{3.2} & \mainmeansd{12.5}{1.3} & \mainmeansd{22.9}{2.0} & \mainmeansd{3.9}{0.8} & \mainmeansd{10.7}{2.4} & \mainmeansd{50.0}{2.4} & \mainmeansd{6.7}{0.1} & \mainmeansd{8.0}{0.7} & \mainmeansd{17.5}{6.3} & \mainmeansd{15.6}{0.4} \\
MPO & \mainmeansd{32.4}{6.6} & \mainmeansd{66.2}{5.2} & \mainmeansd{57.2}{1.3} & \mainmeansd{\textbf{94.0}}{2.6} & \mainmeansd{11.5}{1.4} & \mainmeansd{\textbf{48.2}}{3.5} & \mainmeansd{63.5}{2.7} & \mainmeansd{\underline{42.0}}{12.2} & \mainmeansd{39.9}{8.6} & \mainmeansd{28.6}{10.4} & \mainmeansd{\underline{51.8}}{1.4} \\
AutoRefine$^{\dagger}$ & \mainmeansd{17.5}{3.3} & \mainmeansd{64.2}{9.6} & \mainmeansd{56.7}{1.5} & \mainmeansd{76.8}{0.7} & \mainmeansd{21.0}{4.6} & \mainmeansd{32.7}{5.3} & \mainmeansd{65.6}{3.0} & \mainmeansd{20.7}{10.6} & \mainmeansd{41.1}{2.4} & \mainmeansd{\underline{36.1}}{13.1} & \mainmeansd{46.0}{2.3} \\
SkillOpt$^{\dagger}$ & \mainmeansd{\underline{41.2}}{3.3} & \mainmeansd{\textbf{71.3}}{0.4} & \mainmeansd{\underline{58.1}}{7.4} & \mainmeansd{82.9}{1.1} & \mainmeansd{\underline{25.3}}{5.4} & \mainmeansd{26.6}{8.1} & \mainmeansd{\textbf{73.0}}{2.7} & \mainmeansd{33.4}{11.4} & \mainmeansd{\textbf{46.6}}{6.5} & \mainmeansd{7.6}{6.9} & \mainmeansd{50.7}{1.2} \\
\midrule
Ours  & \mainmeansd{\textbf{58.2}}{4.4} & \mainmeansd{\underline{69.2}}{4.1} & \mainmeansd{\textbf{59.5}}{2.0} & \mainmeansd{\underline{91.9}}{4.7} & \mainmeansd{\textbf{28.8}}{4.6} & \mainmeansd{\underline{43.1}}{7.3} & \mainmeansd{\underline{68.8}}{3.7} & \mainmeansd{\textbf{44.8}}{12.5} & \mainmeansd{\underline{45.1}}{5.1} & \mainmeansd{20.7}{8.5} & \mainmeansd{\textbf{57.1}}{4.0} \\
\midrule
\multicolumn{12}{c}{Qwen3.5-9B~\citep{qwen3.5}} \\
\midrule
ReAct & \mainmeansd{31.1}{1.4} & \mainmeansd{54.2}{8.3} & \mainmeansd{\underline{64.1}}{5.4} & \mainmeansd{90.3}{1.2} & \mainmeansd{\underline{40.6}}{5.3} & \mainmeansd{57.0}{6.4} & \mainmeansd{69.8}{2.7} & \mainmeansd{20.0}{4.6} & \mainmeansd{52.3}{12.4} & \mainmeansd{\textbf{57.0}}{10.2} & \mainmeansd{56.5}{2.1} \\
AdaPlanner & \mainmeansd{34.5}{8.8} & \mainmeansd{24.4}{6.1} & \mainmeansd{46.3}{0.8} & \mainmeansd{44.2}{2.9} & \mainmeansd{18.1}{5.6} & \mainmeansd{19.4}{5.2} & \mainmeansd{35.9}{2.4} & \mainmeansd{4.4}{2.0} & \mainmeansd{28.8}{6.9} & \mainmeansd{48.0}{7.5} & \mainmeansd{32.7}{2.2} \\
MPO & \mainmeansd{25.1}{1.8} & \mainmeansd{56.4}{1.5} & \mainmeansd{\underline{64.1}}{5.8} & \mainmeansd{91.6}{0.9} & \mainmeansd{37.0}{1.9} & \mainmeansd{59.7}{4.8} & \mainmeansd{71.4}{8.2} & \mainmeansd{23.4}{13.5} & \mainmeansd{49.4}{7.5} & \mainmeansd{52.0}{6.9} & \mainmeansd{55.8}{1.5} \\
AutoRefine$^{\dagger}$ & \mainmeansd{28.0}{3.0} & \mainmeansd{50.5}{2.9} & \mainmeansd{59.7}{5.4} & \mainmeansd{89.1}{2.3} & \mainmeansd{36.2}{2.5} & \mainmeansd{58.3}{5.0} & \mainmeansd{68.3}{14.5} & \mainmeansd{19.7}{8.0} & \mainmeansd{58.0}{16.8} & \mainmeansd{\underline{55.5}}{7.8} & \mainmeansd{55.2}{1.2} \\
SkillOpt$^{\dagger}$ & \mainmeansd{\textbf{54.0}}{7.7} & \mainmeansd{\underline{63.0}}{4.2} & \mainmeansd{63.4}{3.0} & \mainmeansd{\underline{91.7}}{6.1} & \mainmeansd{27.1}{0.6} & \mainmeansd{\textbf{64.5}}{7.5} & \mainmeansd{\underline{77.8}}{2.7} & \mainmeansd{\textbf{56.4}}{6.9} & \mainmeansd{\textbf{94.2}}{2.5} & \mainmeansd{45.4}{8.0} & \mainmeansd{\underline{66.8}}{2.1} \\
\midrule
Ours  & \mainmeansd{\underline{50.7}}{3.3} & \mainmeansd{\textbf{70.1}}{4.9} & \mainmeansd{\textbf{82.2}}{4.6} & \mainmeansd{\textbf{96.8}}{2.1} & \mainmeansd{\textbf{42.0}}{6.8} & \mainmeansd{\underline{63.1}}{4.0} & \mainmeansd{\textbf{79.4}}{5.5} & \mainmeansd{\underline{44.1}}{15.7} & \mainmeansd{\underline{82.6}}{7.5} & \mainmeansd{48.9}{10.2} & \mainmeansd{\textbf{69.5}}{2.3} \\
\midrule
\multicolumn{12}{c}{Qwen3.5-27B~\citep{qwen3.5}} \\
\midrule
ReAct & \mainmeansd{77.4}{1.9} & \mainmeansd{68.2}{2.4} & \mainmeansd{75.4}{3.0} & \mainmeansd{\underline{96.7}}{2.3} & \mainmeansd{41.8}{13.3} & \mainmeansd{89.4}{6.9} & \mainmeansd{63.5}{2.7} & \mainmeansd{20.3}{10.6} & \mainmeansd{79.8}{1.5} & \mainmeansd{\underline{70.3}}{15.3} & \mainmeansd{73.6}{1.9} \\
AdaPlanner & \mainmeansd{56.8}{9.6} & \mainmeansd{37.2}{3.4} & \mainmeansd{63.1}{3.6} & \mainmeansd{78.3}{8.8} & \mainmeansd{31.5}{5.9} & \mainmeansd{37.0}{1.5} & \mainmeansd{48.7}{14.1} & \mainmeansd{12.4}{1.5} & \mainmeansd{43.0}{6.4} & \mainmeansd{60.3}{3.1} & \mainmeansd{50.7}{3.6} \\
MPO & \mainmeansd{70.6}{2.3} & \mainmeansd{62.3}{5.0} & \mainmeansd{75.1}{1.5} & \mainmeansd{94.7}{1.2} & \mainmeansd{42.7}{1.1} & \mainmeansd{\underline{98.5}}{2.6} & \mainmeansd{63.5}{2.7} & \mainmeansd{46.1}{2.9} & \mainmeansd{\underline{80.0}}{2.3} & \mainmeansd{64.4}{3.8} & \mainmeansd{73.4}{0.8} \\
AutoRefine$^{\dagger}$ & \mainmeansd{74.4}{5.3} & \mainmeansd{67.4}{5.6} & \mainmeansd{76.5}{3.9} & \mainmeansd{96.6}{1.7} & \mainmeansd{37.5}{3.5} & \mainmeansd{87.0}{1.1} & \mainmeansd{57.1}{4.8} & \mainmeansd{20.0}{13.5} & \mainmeansd{69.9}{13.5} & \mainmeansd{62.5}{2.6} & \mainmeansd{70.4}{0.4} \\
SkillOpt$^{\dagger}$ & \mainmeansd{\textbf{87.5}}{1.9} & \mainmeansd{\textbf{78.3}}{4.3} & \mainmeansd{\underline{77.1}}{0.9} & \mainmeansd{\textbf{100.0}}{0.0} & \mainmeansd{\underline{47.7}}{1.6} & \mainmeansd{\textbf{100.0}}{0.0} & \mainmeansd{\textbf{88.9}}{2.7} & \mainmeansd{\textbf{86.9}}{6.1} & \mainmeansd{\textbf{100.0}}{0.0} & \mainmeansd{65.3}{5.5} & \mainmeansd{\textbf{85.5}}{0.6} \\
\midrule
Ours & \mainmeansd{\underline{84.4}}{3.6} & \mainmeansd{\underline{69.6}}{8.7} & \mainmeansd{\textbf{78.1}}{2.7} & \mainmeansd{94.9}{4.1} & \mainmeansd{\textbf{63.1}}{6.1} & \mainmeansd{\textbf{100.0}}{0.0} & \mainmeansd{\underline{68.3}}{5.5} & \mainmeansd{\underline{59.3}}{6.4} & \mainmeansd{\textbf{100.0}}{0.0} & \mainmeansd{\textbf{76.2}}{1.5} & \mainmeansd{\underline{82.1}}{2.9} \\
\bottomrule
\end{tabular*}

\caption{ScienceWorld final scores (0--100) on 211 test tasks, reported as the mean with sample standard deviation in subscript across three evaluation runs. Avg. weights categories by their task counts within each run. Formatting and $^{\dagger}$ follow Table~\ref{tab:main_results_alfworld}.}
% \vspace{-0.5em}
\label{tab:main_results}
\end{table}

On ScienceWorld, MetaDecomp leads with the smaller actors, whereas SkillOpt leads with the largest actor (Table~\ref{tab:main_results}). With Qwen3.5-9B, MetaDecomp improves over the matched empty-memory control in Mendelian Genetics but remains below ReAct. Plant-growth experience covers shared early steps but not the later requirements of fruit production (Table~\ref{tab:scienceworld-variation}). Such partial guidance may contribute to the low fruit score, illustrating the risk behind the low-performance area studied in Section~\ref{sec:generalization}.

\begin{figure}[!htb]
\centering
\includegraphics[width=0.48\linewidth]{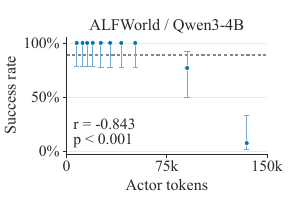}\hfill
\includegraphics[width=0.48\linewidth]{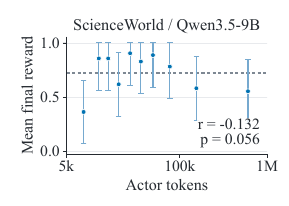}
\caption{\raggedright Performance versus input-plus-output tokens in actor action-generation calls. Each panel has ten bins with 95\% CIs, an overall mean (dashed), and Pearson $r$. ScienceWorld uses a log axis (Appendix~\ref{app:failure-horizon}).}
\label{fig:horizon-analysis}
\end{figure}

\paragraph{Task horizon.}
Figure~\ref{fig:horizon-analysis} measures observed execution length using input-plus-output tokens in actor action-generation calls. ALFWorld success stays high through most bins and falls sharply in the longest executions. On ScienceWorld, mean reward remains relatively stable across intermediate bins and decreases only modestly in the longest executions, whereas full completion drops sharply because intermediate subgoals still earn partial credit when the final goal remains unmet (Appendix~\ref{app:failure-horizon}).

\begin{figure}[!htb]
\centering
\includegraphics[width=\linewidth]{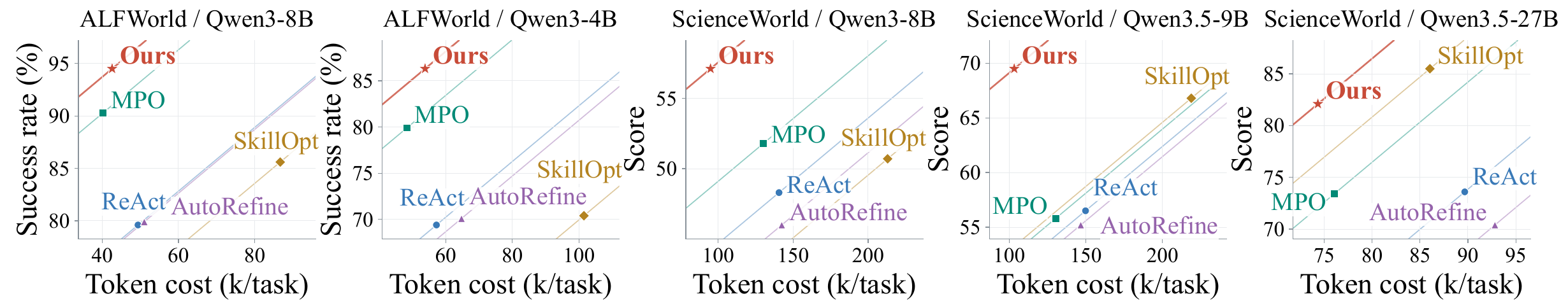}
\caption{Method inference cost and task performance across five settings. Cost is mean input-plus-output tokens per task (thousands), including initial-plan generation, refinement, and execution/control calls in successful and failed episodes. Training is excluded from this figure and reported separately in Table~\ref{tab:learning-inference-cost}; Table~\ref{tab:token_cost} gives the inference input/output breakdown. Axes are scaled independently. Diagonals indicate constant performance minus cost after normalizing each displayed axis range to $[0,1]$. AdaPlanner's performance and costs are reported in Tables~\ref{tab:main_results_alfworld}--\ref{tab:learning-inference-cost}.}
\label{fig:cost-performance}
\end{figure}

\paragraph{Learning and inference cost.}
Figure~\ref{fig:cost-performance} compares performance with inference cost. MetaDecomp achieves higher performance with fewer tokens than ReAct and AutoRefine in all five settings, and improves both measures over SkillOpt in four. Against MPO, ScienceWorld improves in both performance and token use, while ALFWorld trades higher token use for higher success. Costs include initial planning, refinement, actor rollouts, and online control in successful and failed episodes (input/output breakdown: Table~\ref{tab:token_cost}).

Table~\ref{tab:learning-inference-cost} reports full training costs. MetaDecomp uses 30.6--63.1\% fewer training tokens than SkillOpt across all five settings. AutoRefine uses fewer training tokens, while MetaDecomp delivers higher performance at lower inference cost. Only outer-loop learning uses GLM; inner-loop planning and rollouts use the actor model (Table~\ref{tab:training-cost-components}).

\begin{table}[t]
\centering
\renewcommand{\arraystretch}{1.12}
\begin{minipage}[t]{0.485\textwidth}
\centering
{\small\textbf{(a) Training (M tokens)}}\par\vspace{3pt}
\fontsize{8}{9.5}\selectfont
\setlength{\tabcolsep}{1.4pt}
\begin{tabular*}{\linewidth}{@{\extracolsep{\fill}}lccccc@{}}
\toprule
\multirow{2}{*}{Method} & \multicolumn{2}{c}{ALFWorld} & \multicolumn{3}{c}{ScienceWorld} \\
\cmidrule(lr){2-3}\cmidrule(lr){4-6}
 & 4B & 8B & 8B & 9B & 27B \\
\midrule
\textbf{MetaDecomp} & 23.02 & 20.34 & 65.90 & 37.08 & 35.42 \\
ReAct & N/A & N/A & N/A & N/A & N/A \\
AdaPlanner & N/A & N/A & N/A & N/A & N/A \\
MPO & N/A & N/A & N/A & N/A & N/A \\
AutoRefine & 5.29 & 4.08 & 11.35 & 14.81 & 6.89 \\
SkillOpt & 43.62 & 31.44 & 95.00 & 100.51 & 52.97 \\
\bottomrule
\end{tabular*}
\end{minipage}\hfill
\begin{minipage}[t]{0.485\textwidth}
\centering
{\small\textbf{(b) Inference (K tokens/task)}}\par\vspace{3pt}
\fontsize{8}{9.5}\selectfont
\setlength{\tabcolsep}{1.4pt}
\begin{tabular*}{\linewidth}{@{\extracolsep{\fill}}lccccc@{}}
\toprule
\multirow{2}{*}{Method} & \multicolumn{2}{c}{ALFWorld} & \multicolumn{3}{c}{ScienceWorld} \\
\cmidrule(lr){2-3}\cmidrule(lr){4-6}
 & 4B & 8B & 8B & 9B & 27B \\
\midrule
\textbf{MetaDecomp} & 53.80 & 42.60 & 94.90 & 103.30 & 74.30 \\
ReAct & 57.22 & 49.40 & 140.58 & 149.82 & 89.67 \\
AdaPlanner & 32.30 & 28.40 & 40.07 & 42.88 & 30.81 \\
MPO & 48.34 & 40.27 & 130.28 & 130.31 & 76.05 \\
AutoRefine & 64.75 & 51.04 & 142.41 & 146.70 & 92.82 \\
SkillOpt & 101.50 & 86.71 & 213.01 & 218.77 & 86.04 \\
\bottomrule
\end{tabular*}
\end{minipage}
\caption{Total training tokens (left) and mean inference tokens per task (right). Training covers planning, memory learning, and actor rollouts. Inference includes initial-plan generation including MPO initialization, refinement, and all execution/control calls. N/A denotes no training run here. ALFWorld uses Qwen3; ScienceWorld/9B and /27B use Qwen3.5.}
\label{tab:learning-inference-cost}
\end{table}

\subsection{Ablation Study}
\label{sec:loop-ablation}
\label{sec:initialization}
To isolate the contributions of outer-loop learning and inner-loop memory use, Table~\ref{tab:component-ablation} separates learning interventions from inference interventions on ALFWorld and ScienceWorld, where all configurations share cached MPO initial plans and the same model for acting and control. Both retrained variants retain initial refinement and online revision. Fixing the initial schema reduces performance, supporting an adaptive state space. Omitting historical reprojection also reduces performance despite continued schema expansion, supporting reassessment of earlier experience.

At inference, unlinking states and procedures reduces performance while preserving dependencies inside each transition DAG, supporting the use of source and target states to index procedures. Supplying only states or only procedures also lowers performance, confirming that states and procedures are needed together. Removing initial refinement while retaining online memory use likewise lowers performance, separating the benefit of improving the starting plan from later corrections.

\begin{table}[htb]
\centering\small
\setlength{\tabcolsep}{3pt}
\renewcommand{\arraystretch}{1.08}
\begin{tabular*}{0.64\linewidth}{@{\extracolsep{\fill}}lcc@{}}
\toprule
Variant & ALF & SW \\
\midrule
\textbf{Full MetaDecomp} & \textbf{88.81} & \textbf{72.20} \\
\midrule
\multicolumn{3}{@{}l}{\textit{Retrained variants}} \\
Fixed initial schema & 79.10 & 65.02 \\
No historical reprojection & 82.84 & 62.92 \\
\addlinespace[3pt]
\multicolumn{3}{@{}l}{\textit{Inference interventions}} \\
No state--procedure links & 81.34 & 66.29 \\
State only & 80.60 & 53.40 \\
Procedure only & 80.60 & 62.18 \\
No initial refinement & 79.10 & 56.92 \\
\bottomrule
\end{tabular*}
\caption{\raggedright Single-run component ablations: ALFWorld/Qwen3-4B (ALF, success \%) and ScienceWorld/Qwen3.5-9B (SW, score). Retrained variants use the full system's training settings. Removing state--procedure links retains dependencies within each transition DAG.}
\label{tab:component-ablation}
\end{table}

\paragraph{Initial planner.}
Table~\ref{tab:initialization} compares MPO with task-only, ReWOO~\citep{xu2023rewoo}, and ADaPT initialization. Non-MPO plans are generated by the actor model. Test swap replaces the evaluation initializer while retaining MPO-trained memory; Retrain uses the same initializer during learning and testing. Task-only plans from public task information without a cached decomposition; memory remains available. For every alternative, Retrain scores higher than Test swap on both benchmarks. These Retrain results demonstrate that MetaDecomp can learn and operate without MPO's cached decompositions, while the MPO configuration retains the highest scores.

\begin{table}[htb]
\centering\small
\setlength{\tabcolsep}{3pt}
\renewcommand{\arraystretch}{1.08}
\begin{tabular*}{0.60\linewidth}{@{\extracolsep{\fill}}lcccc@{}}
\toprule
\multirow{2}{*}{Initializer} & \multicolumn{2}{c}{Test swap} & \multicolumn{2}{c}{Retrain} \\
\cmidrule(lr){2-3}\cmidrule(l){4-5}
 & ALF & SW & ALF & SW \\
\midrule
MPO & 86.30 & 69.5 & 86.30 & 69.5 \\
Task-only & 77.60 & 57.6 & 85.07 & 65.7 \\
ReWOO & 76.10 & 61.9 & 85.07 & 67.2 \\
ADaPT & 74.60 & 62.1 & 82.09 & 65.7 \\
\bottomrule
\end{tabular*}
\caption{\raggedright Initialization sensitivity: ALFWorld/Qwen3-4B (ALF, success \%) and ScienceWorld/Qwen3.5-9B (SW, score). MPO averages three seeds; alternative ALF entries are single runs on 134 tasks. Protocol: Appendix~\ref{app:component-details}.}
\label{tab:initialization}
\end{table}

\subsection{Generalization, Transfer, and Continual Learning}
\label{sec:generalization}
We first test frozen transfer between ALFWorld task types without additional learning. Table~\ref{tab:cross-type-transfer} shows both gains and losses relative to empty memory, suggesting that transfer depends on workflow coverage: Clean, Cool, Heat, and Pick-Two memories encompass Pick's search--pickup--placement workflow and improve Pick, whereas Pick omits conditioning and lowers Cool and Clean.

Motivated by this observation, we construct a larger, more diverse fusion benchmark of 480 tasks drawn equally from TWC, Coin, Map Reader, Arithmetic, TextCraft, Sorting, BabyAI, and WebShop (Appendix~\ref{app:cross-benchmark-transfer}). Qwen3-4B handles planning and execution with a frozen memory learned across all six ALFWorld types either visible or hidden, without target-domain learning. For analysis, we group tasks by how well the source memory covers their required operations and preconditions. GPT-6.1-sol rates coverage from the source memory, initial public task information, and environment rules, without evaluation outcomes; these ratings define only the analysis groups (Appendix~\ref{app:compatibility-rubric}).

\begin{table}[htb]
\centering\small
\setlength{\tabcolsep}{5pt}
\renewcommand{\arraystretch}{1.08}
\begin{tabular}{@{}lccccccc@{}}
\toprule
Target / source & Look & Pick & Clean & Cool & Heat & Pick-Two & Empty \\
\midrule
Look & 100.00 & 72.22 & 81.48 & 92.59 & 88.89 & 83.33 & 87.04 \\
Pick & 88.89 & 100.00 & 95.83 & 91.67 & 100.00 & 100.00 & 87.50 \\
Clean & 74.19 & 69.89 & 77.42 & 73.12 & 77.42 & 80.65 & 72.04 \\
Cool & 71.43 & 50.79 & 63.49 & 80.95 & 77.78 & 76.19 & 71.43 \\
Heat & 84.06 & 66.67 & 69.57 & 78.26 & 88.41 & 85.51 & 75.36 \\
Pick-Two & 74.51 & 66.67 & 60.78 & 74.51 & 58.82 & 74.51 & 72.55 \\
\bottomrule
\end{tabular}
\caption{ALFWorld cross-type transfer, mean success rate (\%) over three evaluations, with Qwen3-4B for acting, refinement, and online control, and cached MPO initial plans. Rows are target types and columns identify type-specific frozen source memories; Empty uses no memory.}
\label{tab:cross-type-transfer}
\end{table}

Table~\ref{tab:ood-transfer} validates this explanation and identifies a \emph{low-performance area} at partial coverage: relative to hidden memory, visible memory raises success at high coverage, lowers it at partial coverage, and changes little at low coverage, where unmatched states are not recalled. At partial coverage, source methods match early operations but leave remaining steps or preconditions unsupported, steering planning toward incomplete continuations~\citep{dong2026skills}.

\begin{table}[!htb]
\centering\small
\setlength{\tabcolsep}{2.5pt}
\renewcommand{\arraystretch}{1.12}
\begin{tabular*}{0.60\linewidth}{@{\extracolsep{\fill}}lcccc@{}}
\toprule
Coverage & Tasks & Hidden & Visible & $\Delta$ (pp) \\
\midrule
High & 83 & 67.47 & 79.52 & $\mathbf{+12.05}$ \\
Partial & 265 & 70.19 & 67.17 & $\mathbf{-3.02}$ \\
Low & 132 & 27.27 & 28.03 & $+0.76$ \\
\midrule
\textbf{All tasks} & 480 & 57.92 & 58.54 & $+0.63$ \\
\bottomrule
\end{tabular*}
\caption{\raggedright Zero-shot transfer on the eight-benchmark fusion panel: success (\%) with ALFWorld source memory hidden or visible. Operation/precondition coverage defines high (7--10), partial (4--6), and low (1--3) strata. $\Delta$ is Visible minus Hidden, computed from success counts before rounding. Protocol and rubric: Appendix~\ref{app:transfer}.}
\label{tab:ood-transfer}
\end{table}

Continual learning across ALFWorld types follows the same coverage pattern as one shared memory evolves sequentially (Tables~\ref{tab:continual-metrics} and~\ref{tab:continual-matrix}; Appendix~\ref{app:continual}). Through evidence-linked historical reprojection and structural validation, MetaDecomp leads final success, average incremental accuracy, and forward transfer alongside positive backward transfer, while temporary performance drops occur where accumulated procedures only partially match task preconditions or goals.

% \vspace{-0.5em}
\section{Conclusion}
\label{sec:conclusion}
% \vspace{-0.5em}
\leavevmode MetaDecomp learns a shared procedural prior through in-context meta-learning, jointly evolving progress abstractions and reusable procedures to guide task-specific adaptation from execution feedback. It achieves strong task performance with fewer inference tokens than ReAct and AutoRefine. Transfer reveals a low-performance area: partial coverage degrades success relative to hidden memory, whereas high coverage improves success and low coverage changes little. Effective reuse therefore requires matching learned procedures to the task's preconditions and remaining work.

\label{mainbody:end}
\section*{Ethics Statement}
Our experiments use simulated benchmarks, including ALFWorld, ScienceWorld, and the eight-benchmark fusion panel, and do not involve human participants or deployment in physical environments. Procedural memory can propagate incorrect or inappropriate behaviors across tasks. Practical deployment should therefore validate stored procedures, restrict permitted actions, and provide appropriate human oversight.

\section*{Reproducibility Statement}
Section~\ref{sec:method} and Appendix~\ref{app:method} describe the procedural-memory representation and the outer- and inner-loop algorithms. Appendix~\ref{app:implementation} reports model configurations, generation and execution limits, evaluation task counts, the three evaluation seeds, and representative prompt excerpts. Appendix~\ref{app:supplementary} reports training resources, memory size and reuse, component interventions, transfer and continual-learning results, and learning and inference costs. Evaluation protocols accompany the corresponding experiments. The theoretical assumptions and proofs are given in Appendix~\ref{app:theory}. Code will be released upon acceptance.

\section*{AI Use Statement}
We used generative AI tools to improve the language and readability of the manuscript and to assist with code implementation and debugging. The coverage analysis also uses LLM ratings under the protocol in Appendix~\ref{app:compatibility-rubric}. The authors take full responsibility for the final text, code, and reported results, including all AI-assisted contributions.

\label{maintext:end}

\bibliography{main}
\bibliographystyle{iclr2027_conference}

\appendix
\section{Experimental Details}
\label{app:experiments}

\subsection{Implementation Details}
\label{app:implementation}
\label{app:checkpoint-selection}

In the main benchmark evaluations, ScienceWorld uses 211 test instances from 30 native task types, grouped into ten reporting categories. ALFWorld uses 134 unseen instances across six task types. Table~\ref{tab:evaluation-counts} lists their sizes. Evaluation measures transfer to new instances within these task types. Each learning space maintains its own state space and transition DAGs and may cover a native task type or a group of types. For example, ScienceWorld Qwen3.5-9B shares one learning space across all four Matter subtypes.

\begin{table}[htbp]
\centering
\small
\begin{tabular*}{0.7\textwidth}{@{\extracolsep{\fill}}llc@{}}
\toprule
Benchmark & Category & Test tasks \\
\midrule
\multirow[c]{10}{*}{ScienceWorld} & Changes of State (Matter) & 30 \\
& Measurement Instrument (Meas.) & 23 \\
& Electrical Circuits (Elec.) & 25 \\
& Classification (Class.) & 29 \\
& Growing Plants (Plants) & 14 \\
& Chemistry (Chem.) & 22 \\
& Life Spans (Life-1) & 21 \\
& Life Stages (Life-2) & 9 \\
& Forces (Force) & 23 \\
& Mendelian Genetics (Mende.) & 15 \\
\midrule
\multirow[c]{6}{*}{ALFWorld} & Pick \& Place (Pick) & 24 \\
& Clean \& Place (Clean) & 31 \\
& Heat \& Place (Heat) & 23 \\
& Cool \& Place (Cool) & 21 \\
& Examine in Light (Look) & 18 \\
& Pick Two \& Place (Pick-Two) & 17 \\
\bottomrule
\end{tabular*}
\caption{Category sizes for the fixed evaluation panels. Categories use the original benchmark names and follow the main-table order, with the corresponding main-table abbreviations in parentheses. ScienceWorld contains 211 tasks and ALFWorld contains 134 tasks. These counts determine the weights in Avg.}
\label{tab:evaluation-counts}
\end{table}

For ScienceWorld, let $r_i=\max(0,\mathrm{finalScore}_i/100)$ be the normalized score at episode termination, including partial credit. For ALFWorld, $r_i$ is a binary success indicator. With $n_c$ tasks in category $c$, its reported score is $s_c=100n_c^{-1}\sum_{i\in c}r_i$, and the overall score is
\begin{equation}
 \operatorname{Avg.}=\frac{\sum_c n_c s_c}{\sum_c n_c}
 =\frac{100}{N}\sum_{i=1}^{N}r_i.
 \label{eq:experimental-average}
\end{equation}
Missing final scores contribute zero, and episodes with errors remain in the denominator. For three run-level scores $x_1,x_2,x_3$, we report the mean $\bar{x}=\frac{1}{3}\sum_{j=1}^3 x_j$ and sample standard deviation $s=\sqrt{\frac{1}{2}\sum_{j=1}^3(x_j-\bar{x})^2}$. Avg. is computed within each run using Equation~\ref{eq:experimental-average}, then summarized across runs. Scores are displayed to one decimal place; differences are computed before rounding.

For these main experiments, the actor models are Qwen3-8B, Qwen3.5-9B, and Qwen3.5-27B on ScienceWorld, and Qwen3-4B and Qwen3-8B on ALFWorld. Across all MetaDecomp experiments, initial refinement and the online controller use the same model as the actor. The controller performs state projection, target selection, and plan revision. GLM-5.3-Flash is used only for outer-loop memory learning. Transfer and continual learning use Qwen3-4B for acting, refinement, and online control (Appendices~\ref{app:transfer} and~\ref{app:continual}). We follow the original algorithms for all baselines, including AutoRefine's and SkillOpt's learning and validation procedures and their test-time use of learned artifacts. On ALFWorld and ScienceWorld, cached MPO decompositions are generated with the officially released planner checkpoints. AdaPlanner uses the official ALFWorld notebook implementation, adapted to the ScienceWorld interface. AutoRefine and SkillOpt use GLM-5.3-Flash for experience distillation during training and freeze the resulting libraries before testing.

The main evaluations use three execution seeds: 1234, 1235, and 1236.

For the main experiments, MetaDecomp processes batches randomly sampled from the training pool, admitting only complete successes (final normalized reward one) to outer learning. Regular updates use two successful episodes. Only ScienceWorld with Qwen3-8B or Qwen3.5-9B uses a one-success warmup to address low initial success. Appendix~\ref{app:training-audit} summarizes the training pools and learning resources. The state space and transition DAGs remain fixed within each episode and throughout testing, while the task-specific decomposition can be revised using observations. Appendix~\ref{app:method} specifies the update and adaptation procedures.

In comparisons against an empty-memory control (Tables~\ref{tab:cross-type-transfer}, \ref{tab:continual-matrix}, and~\ref{tab:scienceworld-variation}(a)), the empty-memory configuration and full MetaDecomp share MPO's cached decomposition $D^{\mathrm{base}}$ and the same Qwen model for acting and control. The empty-memory configuration uses an empty procedural memory, preserves the decomposition in dependency order without an initial refinement call, and enables revision from execution feedback, whereas full MetaDecomp uses learned knowledge in both initial refinement and online revision, as specified in Appendix~\ref{app:adaptation}.

Each online review recalls at most 3 target states and 3 shortest simple state paths per target. The default per-call output-token limits are 65,536 for the outer learner, 2,048 for the actor, 8,192 for initial plan refinement and the online controller. Actor and controller temperatures are zero. Episodes allow at most 50 interaction steps on ALFWorld or 100 on ScienceWorld, and 12 plan reviews. The execution budget is 1,000,000 input-plus-output tokens per episode for actor and online-review calls. Initial plan refinement and outer-loop learning are outside this stopping budget. Reported costs include initial refinement in both training and inference, and outer-loop learning during training.

The plan reviewer receives the full visible episode on ALFWorld and recent events together with retrieved earlier context on ScienceWorld. State projection and target selection use the full execution graph.

\subsubsection{Prompt excerpts}
\label{app:prompt-examples}

The following excerpts illustrate state-space learning, transition-DAG aggregation, and online revision, with omissions marked by \texttt{[...]}. D0 denotes the cached decomposition $D^{\mathrm{base}}$, D1 the current decomposition, and a stored method is represented by a transition DAG.

\Needspace{4\baselineskip}
\paragraph{State-space learning.}
\begin{Verbatim}[fontsize=\small,breaklines=true,breaksymbolleft={},frame=single,framesep=6pt,rulecolor=\color{black!25}]
Stage 1 - Propose necessary additions to the reusable state schema.

Use the supplied complete task executions jointly. Task descriptions define requirements; public observations define facts. Do not use hidden state, D0/D1, intended effects, or future outcomes as intermediate-state evidence.

[...]

Add a field or value only if all are true:
1. The distinction is supported at identifiable execution boundaries.
2. It changes remaining task-relevant work, applicability of an evidenced reusable method, or task satisfaction.
3. The current schema cannot already represent or derive it.
4. It is reusable across concrete instances and is not merely an action parameter or method-internal condition.
\end{Verbatim}

\Needspace{4\baselineskip}
\paragraph{Transition-DAG aggregation.}
\begin{Verbatim}[fontsize=\small,breaklines=true,breaksymbolleft={},frame=single,framesep=6pt,rulecolor=\color{black!25}]
Stage 4 - Aggregate evidence into methods for one fixed directed state pair.

[...]

Use each group's own supporting/context evidence to establish its executed actions, effects, and dependencies. Cross-group evidence or existing method support may be used only to support reusable decision reasoning, not to supply missing executed actions or effects.

[...]

Require:
- the complete action DAG to support the fixed source-to-target transition;
- every retained action to remain atomic;
- prerequisites of the first action to hold at the source boundary;
- later prerequisites to be established by earlier retained actions when needed.

[...]

Dependencies reference only actions in that method and encode necessary acyclic precedence.
\end{Verbatim}

\Needspace{4\baselineskip}
\paragraph{Online plan revision.}
\begin{Verbatim}[fontsize=\small,breaklines=true,breaksymbolleft={},frame=single,framesep=6pt,rulecolor=\color{black!25}]
Decide whether the current D1 needs repair after the reported stall. Use task_description, actual execution evidence, and current D1. Experience from the procedural memory is optional supporting knowledge, not a goal by itself.

[...]

Only execution evidence establishes that a prerequisite currently holds. If it does not hold, retain suitable earlier D1 steps that establish it, or add the missing work.
\end{Verbatim}

\subsection{Qualitative example of learned procedural memory}
\label{app:qualitative-memory}

We illustrate ALFWorld's Look task type (examining a target object under a desklamp) using a memory snapshot distinct from the evaluated checkpoint in Table~\ref{tab:memory-summary}. It contains 12 retained state nodes and 18 transition DAGs. Its two roles bind the target object and desklamp to task entities. Six binary fields describe observed progress, recording whether the target object has been located, whether it is carried, whether the lamp has been located, whether it is on, whether the agent is at the lamp, and whether the task is satisfied. Location knowledge is distinct from physical presence, and task satisfaction is recorded separately from possession and illumination.

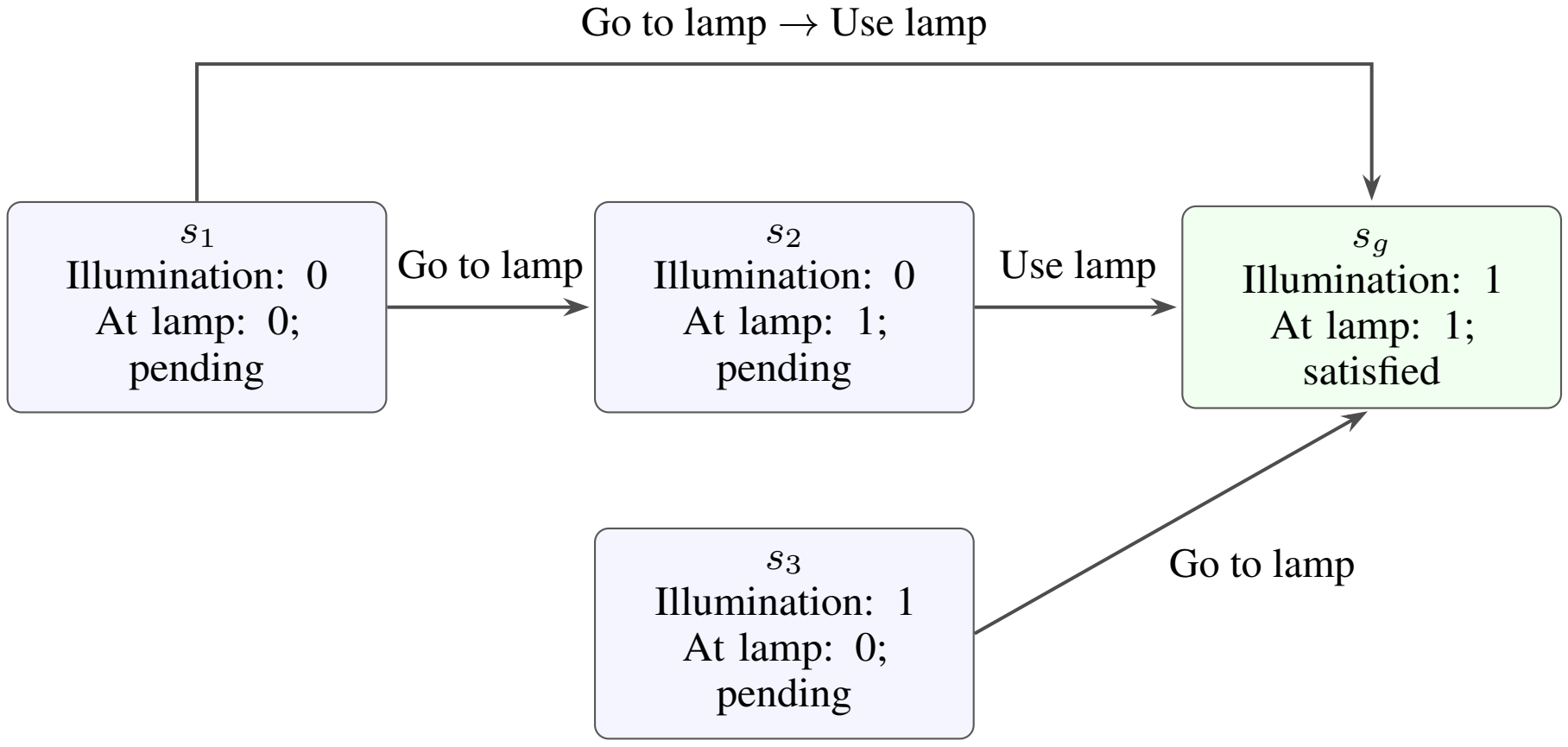
\begin{figure}[htbp]
    \centering
    \begin{tikzpicture}[
    font=\small,
    state/.style={draw=black!65, fill=blue!4, rounded corners=3pt,
        align=center, text width=2.55cm, minimum height=1.25cm, inner sep=5pt},
    procedure/.style={inner sep=1pt, align=center},
    transition/.style={-{Stealth[length=2mm,width=1.4mm]},
        line width=0.75pt, draw=black!70, shorten >=1pt}
]
\node[state] (s1) at (0,0) {$s_1$\\Illumination: 0\\At lamp: 0; pending};
\node[state] (s2) at (4.5,0) {$s_2$\\Illumination: 0\\At lamp: 1; pending};
\node[state, fill=green!6] (sg) at (9,0) {$s_g$\\Illumination: 1\\At lamp: 1; satisfied};
\node[state] (s3) at (4.5,-2.5) {$s_3$\\Illumination: 1\\At lamp: 0; pending};

\draw[transition] (s1.east) -- node[procedure, above=3pt] {Go to lamp} (s2.west);
\draw[transition] (s2.east) -- node[procedure, above=3pt] {Use lamp} (sg.west);
\draw[transition] (s1.north) -- ++(0,1.05) -|
    node[procedure, above=3pt, pos=0.25] {Go to lamp $\rightarrow$ Use lamp} (sg.north);
\draw[transition] (s3.east) -- node[procedure, below right=3pt, pos=0.45] {Go to lamp} (sg.south);
\end{tikzpicture}
    \caption{A subgraph of the learned procedural memory for ALFWorld Look tasks. In all four states, the target object and lamp have been located and the object is carried. Nodes show the remaining three fields. For illumination and presence at the lamp, 1 means currently established by observation and 0 means not established. Each edge is a stored transition DAG, with shortened operation names. The internal arrow in the upper edge's label denotes the dependency between its two operations. State assignments, transition endpoints, and DAG dependencies are preserved from the snapshot.}
    \label{fig:qualitative-memory}
\end{figure}

Figure~\ref{fig:qualitative-memory} illustrates how these distinctions organize the remaining work. In state $s_1$, neither illumination nor presence at the lamp is established by the current evidence. The memory stores both a two-operation DAG that first goes to the lamp's location and then uses it, and separate transitions through $s_2$ for these operations. In $s_3$, illumination is established but the task remains pending. The stored transition reaches $s_g$ by going to the lamp while carrying the object, without an additional use operation. For the two-operation procedure, the actor is instructed to navigate to the receptacle or surface where the lamp was observed, then use the lamp after reaching that location.

The same snapshot also retains alternative procedures between one state pair. From a state where the lamp is located but the target object has not been found, two transition DAGs lead to a state where the object is located but not yet carried. One contains a single operation that repeatedly visits uninspected receptacles until the target is observed. The other orders three operations: visit a receptacle, open it if closed, and visit another receptacle if the object remains unfound. Their shared endpoints preserve a common progress change while their instructions retain different search behavior.

\section{Additional Analyses of Learning, Reuse, and Transfer}
\label{app:supplementary}

\subsection{Training data and resources}
\label{app:training-audit}

For the main experiments, memory learning uses only training trajectories. ALFWorld samples from a candidate pool of 1,200 training instances across six task types (200 per type) and evaluates on 134 unseen test tasks; ScienceWorld uses its official train/validation/test splits of 3,592/1,796/1,819 instances and evaluates on 211 test tasks. The evaluation panels are specified in Appendix~\ref{app:implementation}. Candidate training tasks are sampled randomly in batches without a handcrafted task-type or difficulty curriculum, and the checkpoint with the highest validation performance in each learning space is frozen for testing.

\begin{table}[htbp]
\centering\small
\setlength{\tabcolsep}{5pt}
\begin{tabular}{@{}lccccc@{}}
\toprule
Setting & Spaces & Tasks & Episodes & Successes & Updates \\
\midrule
ALF/Qwen3-4B & 6 & 111 & 111 & 92 & 46 \\
ALF/Qwen3-8B & 6 & 110 & 110 & 94 & 47 \\
SW/Qwen3-8B & 17 & 313 & 330 & 113 & 57 \\
SW/Qwen3.5-9B & 17 & 219 & 228 & 119 & 60 \\
SW/Qwen3.5-27B & 16 & 242 & 242 & 184 & 92 \\
\bottomrule
\end{tabular}
\caption{Training resources for the selected main-experiment checkpoints. Tasks count distinct attempted training tasks, including unsuccessful ones; episodes include retries; successes are complete-success episodes used for learning. Updates count memory updates, including initialization. Counts sum learning spaces within each configuration.}
\label{tab:training-summary}
\end{table}

ALFWorld uses one learning space per task type. For ScienceWorld with Qwen3-8B or Qwen3.5-9B, Matter, Plants, Chemistry, Life Spans, Life Stages, and Forces each share a space across their subtypes. Electrical Circuits groups component-powering tasks separately from conductivity tasks; other native subtypes have separate spaces. Qwen3.5-27B additionally pools subtypes within Measurement and Classification into separate spaces, while separating Chemistry and Forces by native subtype. Supplementary transfer and continual learning use the memories specified in Appendices~\ref{app:transfer} and~\ref{app:continual}.

\subsection{Memory structure and reuse}
\label{app:memory-audit}

Table~\ref{tab:memory-summary} summarizes the learned memory across all five main settings and across individual ALFWorld task types with Qwen3-4B, together with its reuse during test-time planning. Each transition DAG connects a source and target state; alternative procedures sharing the same endpoints capture distinct operation sequences or prerequisite dependencies rather than redundant duplicates. All per-task-type ALFWorld memories remain compact, and in Look every stored transition DAG (9/9) and action node (12/12) is reused at test time.

\begin{table}[htbp]
\centering\small
\setlength{\tabcolsep}{4pt}
\renewcommand{\arraystretch}{1.08}
\begin{tabular}{@{}lccccc@{}}
\toprule
\multicolumn{6}{@{}l}{\textit{(a) Pooled across learning spaces by setting}} \\
Setting & States & DAGs & Actions & DAGs used & Actions used \\
\midrule
ALF/Qwen3-4B & 75 & 106 & 171 & 68/106 & 95/171 \\
ALF/Qwen3-8B & 64 & 89 & 144 & 70/89 & 112/144 \\
SW/Qwen3-8B & 130 & 164 & 359 & 139/164 & 257/359 \\
SW/Qwen3.5-9B & 150 & 161 & 324 & 111/161 & 204/324 \\
SW/Qwen3.5-27B & 163 & 212 & 444 & 140/212 & 295/444 \\
\bottomrule
\end{tabular}

\vspace{6pt}
\begin{tabular}{@{}lccccc@{}}
\toprule
\multicolumn{6}{@{}l}{\textit{(b) By ALFWorld task type with Qwen3-4B}} \\
Task type & States & DAGs & Actions & DAGs used & Actions used \\
\midrule
Look & 7 & 9 & 12 & 9/9 & 12/12 \\
Pick & 15 & 18 & 24 & 15/18 & 21/24 \\
Clean & 16 & 24 & 37 & 15/24 & 21/37 \\
Cool & 6 & 8 & 15 & 5/8 & 9/15 \\
Heat & 20 & 29 & 53 & 9/29 & 11/53 \\
Pick-Two & 11 & 18 & 30 & 15/18 & 21/30 \\
\bottomrule
\end{tabular}
\caption{Memory size and test-time planning reuse across the five main settings (a) and by ALFWorld task type with Qwen3-4B (b). Panel (a) sums independent learning spaces within each configuration; action nodes are counted within each transition DAG. Used items are referenced in an accepted initial plan or an applied revision in at least one test episode, from one evaluation per snapshot including failed episodes.}
\label{tab:memory-summary}
\end{table}

\paragraph{Procedure abstraction.}
The LLM extractor retains an executed operation only if it is judged to advance the source-to-target transition, given the progress already established at the source. During aggregation, an intent is aligned with an existing or new operation, merged with an equivalent operation, or discarded with a reason. Parameterized procedures can be consolidated across differences in concrete objects and the ordering of independent operations.

Test-time planning reuse serves as a proxy for retained-action utility (Table~\ref{tab:memory-summary}). Across the 1,442 stored action nodes in all five configurations, 661/1,442 action nodes (45.84\%) are supported by at least two distinct successful training episodes, 963/1,442 (66.78\%) are reused in at least one test episode, and 746/1,442 (51.73\%) are reused in at least two test episodes.

\subsubsection{Illustrative case: schema expansion and procedure reassignment}
\label{app:schema-reassignment-case}

Consider a task requiring an object to be cleaned and then placed in a receptacle. Suppose an early schema records possession but not cleanliness. Two retained historical continuations can then share the source state ``object held'': one begins before cleanliness is established and contains cleaning followed by placement; the other begins after cleanliness is established and contains placement alone. These procedures can coexist as alternatives under the same coarse endpoint pair.

When cleanliness is added, MetaDecomp revisits the observations available through each retained boundary. The sources become ``held, cleanliness unestablished'' and ``held, cleanliness established''; both successful targets become ``placed, cleanliness established.'' The historical execution spans used to construct each procedure are regrouped by these new source--target pairs. If all of a procedure's spans map to one pair, the procedure can be reassigned there directly; otherwise, it is reassessed separately for each resulting group. The revised memory can now distinguish when cleaning is still required from when placement alone remains.

\subsection{Initialization and component interventions}
\label{app:component-details}

The initialization study uses ALFWorld/Qwen3-4B and ScienceWorld/Qwen3.5-9B (Table~\ref{tab:initialization}). Test swap changes the evaluation initializer while retaining MPO-trained memory; retrain uses the same alternative initializer during learning and testing. Task-only, ReWOO, and ADaPT use the corresponding actor model to generate their initial plans. Task-only initialization constructs a plan from public task information without an external cached decomposition; procedural memory remains available.

For the component study (Table~\ref{tab:component-ablation}), fixed-schema and no-reprojection variants use the full system's training configuration and retain initial refinement and online revision. The former fixes the state-space definition after its initial establishment, while still allowing new state vectors and procedures within that space; the latter omits historical reassessment as the schema evolves. The remaining interventions modify inference: removing state--procedure links retains dependencies within each transition DAG; state-only and procedure-only variants restrict the available knowledge; removing initial refinement retains online memory use.

\subsection{Transfer across task types and benchmarks}
\label{app:transfer}

The within-ALFWorld cross-type study uses Qwen3-4B for acting, initial refinement, and online control, with cached MPO initial plans. GLM-5.3-Flash is used only for source-memory learning. A memory learned from one type is frozen and evaluated on another without target-type learning. Table~\ref{tab:cross-type-transfer} reports means over three evaluations. We observe both positive and negative transfer relative to Empty memory, pointing to the possibility that transfer depends on how completely source procedures cover the target task's required operations and preconditions: Clean, Cool, Heat, and Pick-Two memories already encompass the search--pickup--placement workflow required for Pick and improve Pick above Empty, with Pick-Two also improving Clean and Heat, whereas Pick memory covers searching, picking up, and placing an object directly without intermediate appliance conditioning, steering Cool or Clean tasks to skip cooling or cleaning and place the object prematurely. This observation motivates the larger, more diverse eight-benchmark evaluation below.

\subsubsection{Cross-benchmark evaluation}
\label{app:cross-benchmark-transfer}

The fusion benchmark contains 480 distinct tasks, with 60 each from TextWorld Commonsense (TWC)~\citep{murugesan2021twc}, Coin~\citep{yuan2018counting}, TextCraft~\citep{prasadEtAl2024ADaPT}, BabyAI~\citep{chevalierboisvert2019babyai}, WebShop~\citep{yao2022webshop}, and the TextWorldExpress environments Map Reader, Arithmetic, and Sorting~\citep{jansen2023textworldexpress}. We evaluate full MetaDecomp, including inner-loop adaptation, with Qwen3-4B for initial-plan generation, refinement, online control, and acting (execution seed 42). GLM-5.3-Flash is used only to learn the source memory. The two conditions share the same inference model and differ only in source-memory visibility: one exposes a frozen memory learned jointly across all six ALFWorld task types; the other withholds it. Transfer is zero-shot, with no target-domain learning.

We report binary task success throughout. Within each coverage group, we pool task and success counts and compute differences before rounding. Table~\ref{tab:ood-transfer} reports the resulting success rates and visible-minus-hidden differences.

Relative to hidden memory, success improves at high coverage, declines at partial coverage, and changes little at low coverage, where unmatched states are not recalled. The partial-coverage decline defines the \emph{low-performance area}: when stored procedures match early operations but leave remaining steps or preconditions unsupported, following them steers planning toward incomplete continuations~\citep{dong2026skills}. Appendix~\ref{app:compatibility-rubric} specifies the rating protocol and score groups. Continual learning likewise combines aggregate gains with task-specific regressions as one memory accumulates experience across types (Appendix~\ref{app:continual}).

\paragraph{Task-general applicability.}
MetaDecomp requires no task-specific changes to its learning or adaptation mechanisms. Task instructions, the observation/action interface, and termination and success conditions are specified for each task. State descriptions and procedures are learned from experience, without a hand-written state ontology or procedure library. The transfer evaluations above reuse frozen source memories without target-domain training.

\subsubsection{Operation and precondition coverage rubric}
\label{app:compatibility-rubric}

\paragraph{Rating protocol.}
GPT-6.1-sol, with high reasoning effort, assigns each task an integer operation and precondition coverage score from 1 to 10. It receives the complete source memory, anonymized initial public task content, public environment rules, and the rubric below, without access to execution outcomes. Ratings are finalized before being linked to evaluation outcomes and are used only for analysis, not during agent execution.

\begin{Verbatim}[fontsize=\small,breaklines=true,breaksymbolleft={},frame=single,framesep=6pt,rulecolor=\color{black!25}]
Assessment question
To what extent are specific procedures in the source memory applicable to the target's required operations under the target environment's action semantics and preconditions?

Score anchors
Score 1. Key operations and preconditions are largely uncovered.
Score 5. Some key processes are covered, but important differences in operations or preconditions remain.
Score 10. Nearly all required key operations and applicable preconditions correspond to source methods.

Scoring rules
1. Explain which source procedures apply and which required operations or preconditions lack support, citing evidence for each score deduction.
2. Assess the applicability of parameterized roles and methods, rather than superficial overlap in example names or scene vocabulary.
3. Use only initial public content and public rules. Do not assume hidden facts or turn object attributes into additional goals.
4. Do not deduct points for source methods or state predicates irrelevant to the current task.
5. Public cues and rules provide the application context; their presence alone does not establish that the source memory covers an operation or decision. Distinguish source knowledge from additional reasoning or procedures still required by the target.
6. Do not predict success rates or transfer gains, or adjust scores toward an expected conclusion.
\end{Verbatim}

\paragraph{Coverage groups.}
We group tasks into high coverage (7--10), partial coverage (4--6), and low coverage (1--3).

\subsection{Continual learning and retention}
\label{app:continual}

The shared schema and procedural memory evolve across Look, Pick, Clean, Cool, Heat, and Pick-Two in this order. Each stage uses 20 tasks from ALFWorld's official training split, with updates every four tasks. Qwen3-4B performs acting, initial refinement, and online control; GLM-5.3-Flash is used only for outer-loop memory updates. Each checkpoint is evaluated three times on the official unseen split. Empty is the same agent before memory learning. Test trajectories do not update the memory.

For each method, let $a_{i,j}$ be success on type $i$ after stage $j$, $b_i$ its Empty reference, and $K=6$. We report final average success (Final SR), average incremental accuracy (AIA)~\citep{rebuffi2017icarl}, and forward and backward transfer (FWT and BWT)~\citep{lopezpaz2017gem}. Metrics average task types equally:
\begin{align}
\mathrm{Final\ SR}&=\frac{1}{K}\sum_{i=1}^{K}a_{i,K}, &
\mathrm{AIA}&=\frac{1}{K}\sum_{j=1}^{K}\frac{1}{j}\sum_{i=1}^{j}a_{i,j},\\
\mathrm{FWT}&=\frac{1}{K-1}\sum_{i=2}^{K}(a_{i,i-1}-b_i), &
\mathrm{BWT}&=\frac{1}{K-1}\sum_{i=1}^{K-1}(a_{i,K}-a_{i,i}).
\end{align}
FWT measures performance on a type before its learning stage relative to Empty; BWT measures the final change on earlier types relative to immediately after their respective learning stages. To incorporate new experience while preserving useful procedures as new task types arrive, MetaDecomp expands the state schema and adds new transition DAGs while historical reprojection re-indexes earlier procedures under the updated schema using their saved execution traces, and procedures connecting the same states with different operations or preconditions are retained as separate alternative DAGs rather than overwriting one another; to prevent errors from accumulating in memory, only complete-success episodes update memory, extracted operations must be grounded in execution evidence, and updated transition DAGs are validated for valid node references and acyclicity. MetaDecomp achieves the highest final success, AIA, and FWT, together with positive mean BWT; SkillOpt has the highest BWT (Table~\ref{tab:continual-metrics}).

\begin{table}[htbp]
\centering\small
\setlength{\tabcolsep}{5pt}
\renewcommand{\arraystretch}{1.08}
\begin{tabular}{@{}lcccc@{}}
\toprule
Method & Final SR $\uparrow$ & AIA $\uparrow$ & FWT $\uparrow$ & BWT $\uparrow$ \\
\midrule
MetaDecomp & 82.39 & 86.79 & +0.50 & +0.31 \\
AutoRefine & 70.59 & 75.42 & -0.97 & -0.14 \\
SkillOpt & 80.21 & 85.40 & -12.38 & +5.19 \\
\bottomrule
\end{tabular}
\caption{ALFWorld continual-learning summary. Final SR and AIA are percentages; FWT and BWT are percentage points. All three methods use 20 training tasks per type and update every four tasks. FWT uses each method's own Empty reference. MetaDecomp's Empty reference averages 79.27\%. Definitions and the complete MetaDecomp matrix are in Appendix~\ref{app:continual}.}
\label{tab:continual-metrics}
\end{table}

\begin{table}[htbp]
\centering\small
\setlength{\tabcolsep}{2.5pt}
\renewcommand{\arraystretch}{1.08}
\begin{tabular}{@{}lccccccc@{}}
\toprule
Target & After Look & Pick & Clean & Cool & Heat & Pick-Two & Empty \\
\midrule
Look & 96.30 & 94.44 & 79.63 & 96.30 & 92.59 & 75.93 & 90.74 \\
Pick & 95.83 & 98.61 & 95.83 & 88.89 & 91.67 & 87.50 & 86.11 \\
Clean & 77.42 & 75.27 & 69.89 & 72.04 & 63.44 & 78.49 & 70.97 \\
Cool & 71.43 & 53.97 & 71.43 & 69.84 & 82.54 & 84.13 & 79.37 \\
Heat & 66.67 & 78.26 & 82.61 & 84.06 & 79.71 & 89.86 & 73.91 \\
Pick-Two & 76.47 & 70.59 & 76.47 & 68.63 & 60.78 & 78.43 & 74.51 \\
\bottomrule
\end{tabular}
\caption{MetaDecomp continual learning on ALFWorld's official unseen split (mean success rates over three evaluations, \%). Rows are evaluated types; columns are successive stages of one shared memory. Empty is the reference before memory learning. Qwen3-4B performs acting, refinement, and online control; GLM-5.3-Flash is used only for memory learning.}
\label{tab:continual-matrix}
\end{table}

Inspecting the checkpoint matrix per task type in Table~\ref{tab:continual-matrix} shows that the stage-by-stage results align with the operation and precondition coverage pattern and low-performance area in Table~\ref{tab:ood-transfer}: high coverage improves five of the six task types over Empty at the final stage, whereas partial coverage causes temporary performance drops when accumulated procedures match early search-and-pickup steps but either miss required conditioning steps, such as Cool After Pick (53.97\%), or attach extra steps from later tasks, such as Look After Pick-Two (75.93\%).

\subsection{ScienceWorld variation and task horizon}
\label{app:failure-horizon}

\paragraph{Task difficulty, state abstraction, and procedure reuse.}
Both base-model capabilities and task-dependent sensitivity to state abstraction and procedure reuse contribute to performance variance across ScienceWorld categories and between ScienceWorld and ALFWorld. First, even without learned memory, both ReAct and Empty vary widely across categories with Qwen3.5-9B in Table~\ref{tab:scienceworld-variation}(a), scoring strongly on Classification but dropping on Growing Plants and Life Stages, and ReAct scores lower on ScienceWorld than on ALFWorld at the same Qwen3-8B actor (Tables~\ref{tab:main_results_alfworld} and~\ref{tab:main_results}); scaling the base LLM from Qwen3-8B to Qwen3.5-9B and Qwen3.5-27B substantially raises scores for both ReAct and Full MetaDecomp, lifting categories such as Chemistry and Forces to full completion (Table~\ref{tab:main_results}). Second, isolating state abstraction in Table~\ref{tab:scienceworld-variation}(c) shows that adaptive schemas and historical reprojection yield large gains on multi-stage categories and task types requiring finer progress distinctions, such as Life Stages, Mendelian Genetics, Classification, Pick, and Heat, whereas tasks needing fewer state distinctions, such as Growing Plants or Cool, which required no schema updates during training, change less. Third, isolating state-indexed procedure reuse in Table~\ref{tab:scienceworld-variation}(c) shows strong sensitivity on multi-stage scientific and appliance tasks such as Life Stages, Mendelian Genetics, Cool, and Heat.

\begin{table}[!htb]
\centering\small
\setlength{\tabcolsep}{5pt}
\renewcommand{\arraystretch}{1.10}
\begin{tabular}{@{}lcccc@{}}
\toprule
\multicolumn{5}{@{}l}{\textit{(a) Category scores and full-panel mean}} \\
Category & ReAct & Empty & Full & Full $-$ Empty \\
\midrule
Classification & 90.3 & 88.5 & 96.8 & $+8.3$ \\
Growing Plants & 40.6 & 29.8 & 42.0 & $+12.2$ \\
Life Stages & 20.0 & 22.6 & 44.1 & $+21.5$ \\
Mendelian Genetics & 57.0 & 18.8 & 48.9 & $+30.1$ \\
\midrule
All 211 tasks & 56.5 & 54.6 & 69.5 & $+14.9$ \\
\bottomrule
\end{tabular}

\vspace{6pt}
\begin{tabular}{@{}lcc@{}}
\toprule
\multicolumn{3}{@{}l}{\textit{(b) Plant subtypes sharing one memory}} \\
Subtype & Training successes / episodes & Full test score \\
\midrule
Grow a plant & 6/8 & 68.6 \\
Grow fruit & 0/7 & 15.5 \\
\bottomrule
\end{tabular}

\vspace{6pt}
{\setlength{\tabcolsep}{3.2pt}
\begin{tabular}{@{}lccccccccc@{}}
\toprule
\multicolumn{10}{@{}l}{\textit{(c) Per-category (SW, Qwen3.5-9B) and per-task-type (ALF, Qwen3-4B) component ablations}} \\
Configuration & Class. & Plants & Life-2 & Mende. & SW All & Pick & Cool & Heat & ALF All \\
\midrule
Full MetaDecomp & 97.38 & 49.93 & 46.11 & 60.00 & 72.20 & 100.00 & 80.95 & 91.30 & 88.81 \\
Fixed initial schema & 81.90 & 46.21 & 27.56 & 61.67 & 65.02 & 83.33 & 80.95 & 82.61 & 79.10 \\
No hist.\ reprojection & 93.10 & 45.21 & 15.00 & 20.00 & 62.92 & 87.50 & 85.71 & 82.61 & 82.84 \\
No state--proc.\ links & 95.97 & 39.21 & 26.44 & 42.80 & 66.29 & 100.00 & 61.90 & 82.61 & 81.34 \\
State only & 92.24 & 46.86 & 21.44 & 28.33 & 53.40 & 91.67 & 61.90 & 73.91 & 80.60 \\
Procedure only & 86.48 & 43.21 & 46.44 & 20.00 & 62.18 & 87.50 & 80.95 & 78.26 & 80.60 \\
\bottomrule
\end{tabular}}
\caption{ScienceWorld task variation and component sensitivity. Panel (a) reports three-run means on the 0--100 scale with Qwen3.5-9B; Empty and Full share cached MPO plans and the same actor/control model, and differences are computed before rounding. Panel (b) reports successful and total training episodes for the shared plant memory alongside subtype test scores. Panel (c) reports single-run component ablation scores across representative ScienceWorld categories (Qwen3.5-9B) and ALFWorld task types (Qwen3-4B, success \%).}
\label{tab:scienceworld-variation}
\end{table}

\paragraph{Coverage and remaining scientific requirements.}
The plant subtypes share one memory, but complete-success training episodes cover only growing a plant (Table~\ref{tab:scienceworld-variation}(b)). Fruit production also requires multiple plants, cross-pollination, and fruit development. Plant-growth procedures supply intermediate work, but the memory has no successful fruit-production episode covering the remaining steps. This illustrates the conditions underlying the low-performance area in Section~\ref{sec:generalization}: familiar procedures cover shared steps while leaving later requirements unsupported. Such partial guidance may contribute to the low fruit score. Compared with ALFWorld's recurring object-manipulation routines, ScienceWorld's shared preparation followed by different scientific requirements may expose more tasks to partially applicable guidance.

For the horizon analysis in Figure~\ref{fig:horizon-analysis}, we measure observed execution length by input-plus-output tokens in actor action-generation calls, excluding planning and replanning. Within each benchmark, we pool task types, sort tasks by this measure, and divide them into ten approximately equal-count bins, plotted at their token-interval midpoints. On ALFWorld with Qwen3-4B, success remains high across most bins before dropping to 7.69\% in the longest bin ($r=-0.843$, $p<0.001$). On ScienceWorld with Qwen3.5-9B ($r=-0.132$, $p=0.056$), mean reward remains relatively stable across intermediate bins and decreases only modestly to 0.555 on a 0--1 scale in the longest bin because intermediate subgoals earn partial credit, whereas full task completion drops to 9.52\%.

\subsection{Learning and inference token costs}
\label{app:inference-analysis}

Table~\ref{tab:learning-inference-cost} reports full training totals. Table~\ref{tab:training-cost-components} decomposes MetaDecomp's training into inner-loop planning, outer-loop memory learning, and actor rollouts. Only the outer loop uses GLM-5.3-Flash; inner-loop refinement and online control use the actor model, as do action-generation rollouts. At test time, memory is frozen and only inner-loop planning and rollouts are needed, so outer-loop learning is a one-time training cost that is smaller than SkillOpt across all five settings and is amortized across tasks after deployment. Initial-plan generation, including cached MPO plans, is counted in the inference total.

Table~\ref{tab:token_cost} gives the inference costs underlying Figure~\ref{fig:cost-performance}, including successful and failed episodes from the same three evaluations as the main performance tables. We compute task-count-weighted averages within each run, then report the mean and sample standard deviation across runs. Total cost sums input and output consumption before aggregation.

\providecommand{\mainmeansd}[2]{#1\textsubscript{\tiny$\pm$#2}}
\begin{table}[!htbp]
\centering
\small
\begingroup
\setlength{\tabcolsep}{2pt}
\renewcommand{\arraystretch}{1.06}
\begin{tabular*}{\textwidth}{@{\extracolsep{\fill}}l*{10}{c}@{}}
\toprule
& \multicolumn{6}{c}{ScienceWorld} & \multicolumn{4}{c}{ALFWorld} \\
\cmidrule(lr){2-7}\cmidrule(lr){8-11}
& \multicolumn{2}{c}{Qwen3-8B} & \multicolumn{2}{c}{Qwen3.5-9B} & \multicolumn{2}{c}{Qwen3.5-27B} & \multicolumn{2}{c}{Qwen3-8B} & \multicolumn{2}{c}{Qwen3-4B} \\
\cmidrule(lr){2-3}\cmidrule(lr){4-5}\cmidrule(lr){6-7}\cmidrule(lr){8-9}\cmidrule(lr){10-11}
Method & Input & Output & Input & Output & Input & Output & Input & Output & Input & Output \\
\midrule
ReAct & \mainmeansd{138.7}{6.0} & \mainmeansd{1.8}{0.1} & \mainmeansd{143.9}{2.0} & \mainmeansd{6.0}{0.2} & \mainmeansd{86.4}{3.6} & \mainmeansd{3.3}{0.2} & \mainmeansd{47.8}{0.6} & \mainmeansd{1.6}{0.0} & \mainmeansd{56.5}{1.0} & \mainmeansd{0.7}{0.0} \\
AdaPlanner & \mainmeansd{35.0}{0.8} & \mainmeansd{5.1}{0.3} & \mainmeansd{35.4}{0.7} & \mainmeansd{7.5}{0.1} & \mainmeansd{25.3}{0.4} & \mainmeansd{5.5}{0.1} & \mainmeansd{20.3}{0.2} & \mainmeansd{8.1}{0.1} & \mainmeansd{23.6}{0.4} & \mainmeansd{8.7}{0.1} \\
MPO & \mainmeansd{128.6}{5.0} & \mainmeansd{1.7}{0.1} & \mainmeansd{125.4}{11.8} & \mainmeansd{4.9}{0.4} & \mainmeansd{73.3}{1.2} & \mainmeansd{2.7}{0.0} & \mainmeansd{39.1}{1.3} & \mainmeansd{1.2}{0.1} & \mainmeansd{47.8}{2.3} & \mainmeansd{0.6}{0.0} \\
AutoRefine & \mainmeansd{140.6}{4.4} & \mainmeansd{1.8}{0.1} & \mainmeansd{141.0}{2.0} & \mainmeansd{5.7}{0.1} & \mainmeansd{89.4}{17.4} & \mainmeansd{3.4}{0.7} & \mainmeansd{49.4}{1.2} & \mainmeansd{1.6}{0.0} & \mainmeansd{64.2}{3.4} & \mainmeansd{0.6}{0.0} \\
SkillOpt & \mainmeansd{211.5}{9.3} & \mainmeansd{1.5}{0.1} & \mainmeansd{216.1}{6.8} & \mainmeansd{2.7}{0.1} & \mainmeansd{84.7}{1.4} & \mainmeansd{1.3}{0.0} & \mainmeansd{85.4}{1.4} & \mainmeansd{1.3}{0.0} & \mainmeansd{101.0}{2.8} & \mainmeansd{0.5}{0.0} \\
\midrule
Ours & \mainmeansd{92.8}{1.3} & \mainmeansd{2.1}{0.0} & \mainmeansd{98.6}{3.8} & \mainmeansd{4.7}{0.7} & \mainmeansd{71.1}{5.2} & \mainmeansd{3.2}{0.2} & \mainmeansd{41.2}{0.4} & \mainmeansd{1.4}{0.0} & \mainmeansd{52.7}{2.1} & \mainmeansd{1.1}{0.0} \\
\bottomrule
\end{tabular*}
\endgroup
\caption{Inference input and output tokens per task, in thousands, including initial-plan generation, refinement, and execution/control calls. Values report the mean with sample standard deviation in subscript across the three evaluations used in the main performance tables. Appendix~\ref{app:inference-analysis} gives the aggregation protocol.}
\label{tab:token_cost}
\end{table}

\begin{table}[!htbp]
\centering\small
\renewcommand{\arraystretch}{1.08}
\setlength{\tabcolsep}{7pt}
\begin{tabular}{@{}lcccc@{}}
\toprule
\multirow{2}{*}{Setting} & \multicolumn{2}{c}{Planner} & Actor & \multirow{2}{*}{Total} \\
\cmidrule(lr){2-3}
 & Inner loop & Outer loop & Rollouts & \\
\midrule
ALF/Qwen3-4B & 1.29 & 10.47 & 11.26 & 23.02 \\
ALF/Qwen3-8B & 2.68 & 7.45 & 10.21 & 20.34 \\
SW/Qwen3-8B & 15.74 & 7.60 & 42.56 & 65.90 \\
SW/Qwen3.5-9B & 5.60 & 4.64 & 26.84 & 37.08 \\
SW/Qwen3.5-27B & 2.85 & 10.50 & 22.07 & 35.42 \\
\bottomrule
\end{tabular}
\caption{MetaDecomp training token breakdown (millions). Inner-loop planning and actor rollouts use the listed Qwen model; only outer-loop memory learning uses GLM-5.3-Flash. The three components sum to the full training totals in Table~\ref{tab:learning-inference-cost}.}
\label{tab:training-cost-components}
\end{table}

\section{Method Details}
\label{app:method}

\subsection{Representation and overall algorithm}
\label{app:method-representation}

\subsubsection{State space, transition DAGs, and procedural memory}

Each learning space covers a supplied task type or a predefined group of related types within one benchmark. Its learned schema is
\begin{equation}
 \Sigma=\bigl(\mathcal R,\mathcal F,\{V_f,\iota_f,\sigma_f\}_{f\in\mathcal F}\bigr),
 \label{eq:full-schema}
\end{equation}
where $\mathcal R$ specifies task-relative roles and $\mathcal F$ is a finite set of progress fields. Each field $f$ has finite values $V_f$, an initial value $\iota_f\in V_f$, and an observable interpretation $\sigma_f$. The schema defines $\mathcal X_\Sigma=\prod_{f\in\mathcal F}V_f$. Only retained assignments become state nodes in the memory, so the Cartesian product can include unsupported or unreachable combinations.

A state $s\in\mathcal X_\Sigma$ assigns one declared value to every field. The initial assignment $s_{\mathrm{init}}=(\iota_f)_f$ marks progress not yet established. An unestablished condition need not be physically false. Current-condition fields track observed changes, while retained-knowledge fields follow their declared persistence conditions.

Role bindings $\beta$ associate the referents in $\mathcal R$ with task entities and may remain unresolved. Complete field assignments determine state identity independently of these bindings.

The procedural memory $\mathcal H=(\mathcal S,\mathcal T)$ consists of retained state nodes $\mathcal S\subseteq\mathcal X_\Sigma$ and transition DAGs. Each transition in $\mathcal T$ records a source state, a target state, and a transition DAG $M_e$,
\[
 e=(s_e^-,s_e^+,M_e),\qquad
 M_e=(A_e,E_e,x_e),\qquad s_e^-\ne s_e^+.
\]
Here $A_e$ contains parameterized operations, $E_e$ encodes acyclic prerequisite dependencies inferred from execution evidence, and $x_e(a)$ gives the operation's instruction, including supported applicability, argument-selection, and stopping conditions. Alternative procedures may connect the same state pair. Thus $\mathcal H$ is a directed state multigraph whose state graph can contain cycles, although each transition DAG is acyclic. The dependencies guide planning rather than define an environment transition model.

\subsubsection{Execution evidence and the two-loop cycle}

The actor produces history $h_t=(o_0,a_1,o_1,\ldots,a_t,o_t)$, retained as an execution DAG of actual actions, observations, and execution order, together with $o_0$. This graph supplies evidence for state projection and learning. In the main configuration, the actor is guided by an ordered decomposition $D_t=(d_{t,1},\ldots,d_{t,n_t})$, initialized from a cached baseline DAG $D^{\mathrm{base}}$. Planned instructions do not establish attained progress.

A boundary occurrence $\omega$ identifies the entry observation or the observation after an executed action. A local realization $\ell$ records boundaries $\omega_\ell^-,\omega_\ell^+$, their state assignments and role bindings, and evidence-supported operations proposed to connect their states. Occurrences remain distinct even when their assignments coincide. The evidence ledger $\mathcal P_b$ retains original executions, realizations, state assignments, and mappings from local operations to methods for subsequent reassessment.

Algorithm~\ref{alg:metadecomp} holds $(\Sigma_b,\mathcal H_b)$ fixed within each episode while adapting $D_t$. Successful training executions enter a buffer for their learning space. Outer updates use batches of $B$ successes together with the ledger; an optional one-success warmup initializes learning in the settings specified in Appendix~\ref{app:implementation}. Before the first update, the schema is unavailable and the procedural memory is empty, so initialization uses the cached baseline while online revision uses observations. Evaluation freezes one snapshot $(\Sigma^*,\mathcal H^*)$ per learning space and excludes test executions from the training buffer.

\begin{algorithm}[t]
\caption{MetaDecomp outer-loop learning and inner-loop adaptation}
\label{alg:metadecomp}
\begin{algorithmic}[1]
\REQUIRE Training tasks and cached plans for one learning space, modules $\phi$, batch size $B$
\STATE Initialize $\Sigma$ as unavailable, with empty procedural memory $\mathcal H$, ledger $\mathcal P$, and buffer $\mathcal B$
\FOR{each training task $\tau$}
    \STATE $D\leftarrow\operatorname{Plan}_\phi(\tau,D^{\mathrm{base}};\Sigma,\mathcal H)$, preserving the cached plan if no methods exist
    \STATE Initialize $h\leftarrow(o_0)$ and hold $(\Sigma,\mathcal H)$ fixed throughout the episode
    \WHILE{execution budget remains and the episode is active}
        \STATE Execute an actor action guided by $D$ and visible history
        \STATE Append the action and its observation to $h$
        \IF{execution feedback triggers review}
            \STATE Project $h$ into $\Sigma$ when a schema is available
            \STATE Recall methods on paths from an exact stored-state match, if one exists
            \STATE Revise or retain $D$ using execution evidence and any grounded methods
        \ENDIF
    \ENDWHILE
    \IF{the training task succeeds}
        \STATE $\mathcal B\leftarrow\mathcal B\cup\{(\tau,h)\}$
    \ENDIF
    \IF{$|\mathcal B|=B$ or (warmup enabled and $\Sigma$ unavailable and $|\mathcal B|=1$)}
        \STATE Learn or extend schema $\Sigma$ from $\mathcal B$ to define the state space
        \STATE If the schema changes, reproject historical boundaries using $\mathcal P$
        \STATE Extract observed boundaries, state changes, and supported operations from $\mathcal B$
        \STATE Group new and reassessed realizations by complete source--target assignments
        \STATE Migrate unambiguous old method associations and reassess split support
        \STATE Consolidate supported transition DAGs in $\mathcal H$, retaining alternative methods
        \STATE Publish updated $(\Sigma,\mathcal H,\mathcal P)$ and clear $\mathcal B$
    \ENDIF
\ENDFOR
\STATE For evaluation, freeze $(\Sigma^*,\mathcal H^*)$ and run only within-task planning and execution
\end{algorithmic}
\end{algorithm}

\subsection{Outer-loop learning of the state space and transition DAGs}
\label{app:learning}

An outer update maps the current snapshot, ledger, and batch $\mathcal B_b$ of successful training executions to $(\Sigma_{b+1},\mathcal H_{b+1},\mathcal P_{b+1})$. Separate model calls propose schema additions, extract local realizations, reproject retained historical boundaries, and aggregate methods. The program matches states, reconciles method support, and materializes the procedural memory between these calls.

\subsubsection{Learning the state space and extracting evidence}

The schema learner receives the current schema, task descriptions, and execution graphs in $\mathcal B_b$. It proposes observable distinctions that change remaining work, procedure applicability, or task satisfaction, transfer across instances, and are not already expressible. Independent dimensions occupy separate fields, while values of one field answer the same question. Task constants, concrete identities, temporary search locations, and unsupported values are excluded.

Schema updates may add roles, fields, or field values, preserving existing definitions, meanings, and initial values. Additive vocabulary changes can still revise historical assignments and need not satisfy the refinement assumption in Appendix~\ref{app:theory}.

Under the updated schema, extraction assigns the entry boundary and selects later boundaries at observed state changes used by subsequent operations or marking task completion. Each retained boundary receives a complete field assignment supported by observations available through that boundary. Task descriptions specify requirements and help bind roles, but do not establish attained progress. The extractor proposes transitions between adjacent retained boundaries on the execution path.

Each realization $\ell$ contains operation intents and inferred dependencies. An intent describes one executed atomic operation or its repetition with supported argument-selection and stopping conditions. Distinct operations remain separate intents, parameterized by task roles or relations supported by the task, state, and observations. Failed actions may provide context; invalid attempts and unexecuted plan steps cannot supply reusable operation evidence.

A nonentry boundary $\omega$ is the observation immediately after action $j_\tau(\omega)$. Let $\preceq_\tau$ denote equality or reachability in the execution DAG. Each retained local operation $a$ cites a nonempty set of supporting actions satisfying
\begin{equation}
 \emptyset\ne C_\ell(a)\subseteq
 \{j:j_\tau(\omega_\ell^-)\preceq_\tau j
             \preceq_\tau j_\tau(\omega_\ell^+)\}.
 \label{eq:support-interval}
\end{equation}
For an entry source, the lower bound is omitted. Equation~\ref{eq:support-interval} bounds admissible evidence citations and includes the source action. Extraction and aggregation are instructed to retain operations needed from the already attained source state, distinguishing them from actions that established it. The interval check does not itself enforce this semantic distinction. Context references identify observations but do not count as execution support. Realizations lacking a supported complete transition may be discarded.

\subsubsection{Reprojecting historical boundaries}

When $\Sigma$ changes, the learner reprojects retained historical boundaries from their original execution graphs. Let $\mathcal F_\Delta$ contain new fields and fields with expanded value sets. Remapping has the form
\begin{equation}
 q_{b+1}(\omega)_f=
 \begin{cases}
  \widehat q_{b+1}(\omega)_f,&f\in\mathcal F_\Delta,\\
  q_b(\omega)_f,&f\in\mathcal F_b\setminus\mathcal F_\Delta,
 \end{cases}
 \label{eq:historical-remap}
\end{equation}
where $\widehat q_{b+1}$ is the model's reassessment. Only new role bindings are added. Existing bindings, unaffected field values, boundary locations, and action evidence remain fixed.

Remapping is instructed to use only observations available through each boundary, accounting for intervening changes and declared persistence semantics.

New and reassessed realizations are grouped by exact equality of their complete source and target assignments, excluding role bindings, as in Equation~\ref{eq:experience-groups}.

\subsubsection{Reconciling support and updating transition DAGs}

Let $\psi(\omega)$ be a boundary's reassigned state identity, or $\bot$ for an excluded occurrence. For the complete support set $\operatorname{supp}(e)$ of an old transition $e$, its projected endpoint pairs are
\begin{equation}
 \Pi_b(e)=\{(\psi(\omega_\ell^-),\psi(\omega_\ell^+)):
                     \ell\in\operatorname{supp}(e)\}.
 \label{eq:method-migration}
\end{equation}
A method migrates directly only if $\Pi_b(e)=\{(u,v)\}$ with $u,v\ne\bot$ and $u\ne v$. Split support supplies the old procedure as a candidate for reassessment at valid supported pairs, rather than copying to all endpoint combinations. Empty support, excluded endpoints, or coincident endpoints prevent direct migration. Changed pairs without a migrated method also enter review.

For each pair $(u,v)$ under review, aggregation receives local realizations, supporting observations, and admissible existing methods. It groups procedures by operations, applicability, argument-selection and stopping semantics, and inferred dependencies, ignoring wording, concrete bindings, and ordering of independent operations. Evidence from another realization can inform reusable decision rules but cannot supply missing actions or effects for the current realization.

Let $I_{uv}$ be the local operation intents at this pair, indexed by their realizations, and let $A_m^+$ be the resulting operations of a retained or proposed method $m$. Aggregation supplies the mapping
\begin{equation}
 \mu:I_{uv}\longrightarrow
 \bigcup_m\bigl(\{m\}\times A_m^+\bigr)\cup\{\bot\},
 \label{eq:intent-mapping}
\end{equation}
where $\bot$ denotes a discarded intent. Every intent is accounted for exactly once, all retained intents from one realization select the same method, and every new operation receives mapped execution evidence. Multiple intents may support one operation. Mappings and observations are retained in the ledger.

Aggregation refines supported operations and dependencies while leaving unselected methods unchanged. Dependencies remain acyclic and are transitively reduced. Distinct alternatives remain separate transition DAGs.

Newly extracted states are admitted only when used by a materialized method. Previously retained historical occurrences remain represented under their reassigned states. The update produces the revised schema, procedural memory, and evidence ledger. Representative prompts appear in Appendix~\ref{app:prompt-examples}.

\subsection{Inner-loop adaptation with procedural memory}
\label{app:adaptation}

The inner loop constructs and revises an ordered decomposition while holding $\Sigma$ and $\mathcal H=(\mathcal S,\mathcal T)$ fixed. State projection uses actor observations as evidence of attained progress; planned instructions and retrieved states do not establish it.

\subsubsection{Initial decomposition from a cached baseline}

In the main configuration, initial refinement aligns MPO's cached decomposition $D^{\mathrm{base}}$ with the learned state space and procedures. The baseline DAG is first ordered by dependency. The refinement model receives this sequence, the task, the schema, and all methods in the learning space, including their endpoints, instructions, and precedence edges. Initialization requires neither a stored-state match nor path selection.

Refinement can replace an instruction with one or more ordered steps, insert steps before or after it, or remove it. The resulting nonempty decomposition preserves the baseline order of unchanged instructions and the proposed order of inserted steps.

Grounding binds roles to the current task's entities, destination, and requirements. The model may combine partial methods, repeat operations, or add unstored work. Source states specify prerequisites rather than attained facts, and earlier planned steps must establish required progress. Reused operations must respect their dependencies, and repetition requires an observable stopping condition.

If the procedural memory contains no methods, initialization retains the ordered baseline.

\subsubsection{Online state projection and method recall}

When feedback triggers review, the state projector receives task requirements, the full execution graph, and the previous confirmed projection. Its prompt processes evidence in execution order, separates requirements from attained facts, and gives observations precedence over the previous projection. Ineffective actions alone do not revoke established progress.

A projection is complete when every field has a supported declared value, including any declared value for unknown or unestablished progress. If any field cannot be assigned such a value, the projection is incomplete and is omitted from review. Role bindings may remain unresolved.

A complete projection matches $\mathcal S$ by exact field equality, independently of role bindings. Without a match, the reviewer receives the projection but no recalled methods. Given a match $\widehat s_t$, target selection chooses from reachable stored states using the task, decomposition, execution graph, and stall information. It may select no targets. Retrieved methods are subsequently grounded in the current task and observations.

Let $\mathcal G_t$ denote the selected targets, with $|\mathcal G_t|\leq G_{\max}$, and let $\mathcal P_K(\widehat s_t,g)$ contain up to $K$ shortest simple state paths from $\widehat s_t$ to $g$. For the set $\mathcal M(u,v)$ of alternative methods stored on a state pair, recall returns
\begin{equation}
 \mathcal R_t=\bigcup_{g\in\mathcal G_t}\ \bigcup_{p\in\mathcal P_K(\widehat s_t,g)}
 \ \bigcup_{(u,v)\in\operatorname{edges}(p)}
 \{(u,v,M):M\in\mathcal M(u,v)\}.
 \label{eq:path-retrieval}
\end{equation}
Paths are ranked by transition count, not operation count or execution cost. Each is retained in full with every alternative transition DAG on its state pairs, without a maximum path-length cutoff. Appendix~\ref{app:implementation} specifies $G_{\max}$ and $K$. Review can proceed without recalled methods.

\subsubsection{Grounded revision and continued execution}

The reviewer receives $D_t$, $\Sigma$, $\mathcal R_t$, and the context $c_t$, which includes execution evidence, stall information, any complete current projection, and the cached baseline when available. It infers completed and remaining work from observations and may retain an adequate plan.

Recalled target states are outcomes to attain, while source states specify prerequisites. Partial reuse requires omitted prerequisites to hold in the observed execution or be established by earlier instructions. Otherwise the reviewer adds or preserves the necessary work. Revision may also proceed without reusing a method.

Online revision uses only existing instructions as anchors. Each edit inserts one step before or after an anchor, or replaces that anchor with one step. For the lists $\mathbf u_i$ and $\mathbf v_i$ inserted before and after $d_{t,i}$, and its replacement or unchanged instruction $\widetilde d_{t,i}$, materialization is
\begin{equation}
 D_t^+=\mathop{\Vert}_{i=1}^{n_t}
        \bigl(\mathbf u_i\Vert[\widetilde d_{t,i}]\Vert\mathbf v_i\bigr),
 \label{eq:inner-patch}
\end{equation}
where $\Vert$ denotes concatenation. Each anchor permits at most one replacement, insertions at the same position preserve proposal order, and unmentioned instructions remain. Online edits do not delete, explicitly reorder, or replace an anchor by multiple steps.

The actor resumes with the current decomposition, as in Equation~\ref{eq:adapt-and-act}. Later observations may trigger further reviews while $\Sigma$ and $\mathcal H$ remain fixed.

\subsection{State-abstraction analysis and proofs}
\label{app:theory}

The schema $\Sigma$ defines a state space whose retained nodes and transition DAGs form the procedural memory $\mathcal H=(\mathcal S,\mathcal T)$. We study two simplified questions to motivate this design. The first concerns procedure selection with fixed methods, continuation losses, and a history distribution. The second concerns connections between fixed historical occurrences when their endpoint assignments are refined. These comparisons characterize the value of retained information and the structural effects of refinement under the stated assumptions.

\subsubsection{Decision setting and value of refinement}

Fix one learning space and a nonempty finite dictionary $\mathcal M$ of reusable methods. We compare selectors choosing among the same methods under the same continuation procedure. Let $(H,C)$ have a fixed joint distribution, where $H$ is the observed action--observation history and $C$ contains the task information and role bindings available to every selector. The random history $H$ is distinct from the procedural memory $\mathcal H$. All random variables are defined on a common probability space, and the history and context spaces are assumed to be standard Borel spaces so that the conditional quantities below admit measurable versions. All equalities between conditional quantities are understood almost surely.

For each $m\in\mathcal M$, let $\ell(m,h,c)\in[0,1]$ be a measurable conditional expected continuation loss after grounding and choosing $m$ at $(h,c)$. The continuation procedure is fixed across the comparison. Subsequent environment and execution randomness is integrated into this loss, which can reflect failure, normalized execution cost, or their fixed combination. Any learned objects needed to specify the dictionary or continuation procedure are held fixed. The analysis therefore compares what can be achieved with different information about the same procedural decision.

A measurable projection $q$ maps the history and common context to a complete schema state $S=q(H,C)\in\mathcal X_\Sigma$. A state-limited selector observes $(S,C)$, whereas a history selector observes $(H,C)$. Both receive the same $C$, so the comparison concerns the history information retained by $q$ rather than a difference in task requirements or bindings. Define
\begin{equation}
 \mathfrak R(q)=\mathbb E\!\left[\min_{m\in\mathcal M}
    \mathbb E[\ell(m,H,C)\mid S,C]\right],\qquad
 \mathfrak R_H=\mathbb E\!\left[\min_{m\in\mathcal M}\ell(m,H,C)\right].
 \label{eq:interface-risk}
\end{equation}
Boundedness of the losses makes both risks finite. We next verify that they are the optimal risks for their respective information sets, including when selection may be randomized.

Write
\[
 r_m(S,C)=\mathbb E[\ell(m,H,C)\mid S,C],\qquad
 m_q^*(S,C)\in\operatorname*{arg\,min}_{m\in\mathcal M}r_m(S,C).
\]
Fix an ordering of the finite dictionary and choose the first minimizer whenever there is a tie. This gives a measurable deterministic selector because it requires only finitely many comparisons of measurable functions. A randomized state-limited selector is a probability vector $a(\cdot\mid S,C)$ measurable with respect to $(S,C)$. Its internal randomization is conditionally independent of $H$ given $(S,C)$, so it cannot convey additional history information. Its risk is
\[
 \begin{aligned}
 \mathfrak R(a;q)
 &=\mathbb E\!\left[\sum_{m\in\mathcal M}
       a(m\mid S,C)\ell(m,H,C)\right]\\
 &=\mathbb E\!\left[\sum_{m\in\mathcal M}
       a(m\mid S,C)r_m(S,C)\right]\\
 &\geq\mathbb E\!\left[\min_{m\in\mathcal M}r_m(S,C)\right]
 =\mathfrak R(q).
 \end{aligned}
\]
The second equality conditions on $(S,C)$ and uses the measurability of the selection probabilities. The inequality holds pointwise because the conditional risk is a convex combination of the $r_m$. Choosing $m_q^*$ attains equality, establishing $\inf_a\mathfrak R(a;q)=\mathfrak R(q)$. Thus randomization cannot lower this optimal risk. The same finite-minimum argument applied to $\ell(m,H,C)$ gives a measurable history selector attaining $\mathfrak R_H$. The history selector has access only to observations already obtained, so partial observability or unavoidable continuation loss can still leave $\mathfrak R_H>0$.

Let $q'$ be another measurable projection and write $S'=q'(H,C)$. We say that $q'$ refines $q$ when there is a measurable map $\pi$ such that $q=\pi\circ q'$. In particular, the old state can be recovered from the new state without observing anything beyond it. Adding fields while preserving every old coordinate satisfies this condition by taking $\pi$ to discard the new coordinates. This is a condition on the assignments induced by the projections, not merely on the names or number of schema entries.

\begin{proposition}[Decision value of a state distinction]
\label{prop:decision-interface}
Suppose $q=\pi\circ q'$ and keep the dictionary $\mathcal M$, continuation losses, and joint distribution of $(H,C)$ fixed. Let $m_q^*$ be an optimal selector using $(S,C)$ and let
$r'_m(S',C)=\mathbb E[\ell(m,H,C)\mid S',C]$.
Then
\begin{equation}
 \begin{aligned}
 &\mathfrak R_H\leq\mathfrak R(q')\leq\mathfrak R(q),\\
 &\mathfrak R(q)-\mathfrak R(q')=
 \mathbb E\!\left[r'_{m_q^*(S,C)}(S',C)
             -\min_{m\in\mathcal M}r'_m(S',C)\right].
 \end{aligned}
 \label{eq:decision-refinement}
\end{equation}
The refinement has strictly positive value if and only if the coarse optimal selector is suboptimal under the refined information with positive probability.
\end{proposition}

\begin{proof}
Since $S=\pi(S')$, the information in $(S,C)$ is contained in that in $(S',C)$. For each fixed $m$, the tower property gives
\[
 r_m(S,C)=\mathbb E[r'_m(S',C)\mid S,C].
\]
The coarse selector $m_q^*(S,C)$ is also measurable with respect to $(S',C)$. Expanding its loss over the finite dictionary and conditioning on the refined information therefore yields
\[
 \begin{aligned}
 \mathfrak R(q)
 &=\mathbb E[\ell(m_q^*(S,C),H,C)]\\
 &=\mathbb E\!\left[\sum_{m\in\mathcal M}
       \mathbf 1\{m_q^*(S,C)=m\}\ell(m,H,C)\right]\\
 &=\mathbb E\!\left[\sum_{m\in\mathcal M}
       \mathbf 1\{m_q^*(S,C)=m\}r'_m(S',C)\right]\\
 &=\mathbb E[r'_{m_q^*(S,C)}(S',C)].
 \end{aligned}
\]
Subtracting $\mathfrak R(q')=\mathbb E[\min_m r'_m(S',C)]$ proves the exact improvement identity in Equation~\ref{eq:decision-refinement}. Its integrand is nonnegative, proving $\mathfrak R(q')\leq\mathfrak R(q)$.

For the other inequality, any selector based on $(S',C)$ can be implemented from $(H,C)$ by first applying $q'$. The class of history selectors therefore includes the class of refined-state selectors, giving $\mathfrak R_H\leq\mathfrak R(q')$. Equivalently, the pointwise minimum over history-conditioned losses is no larger than the loss of the refined-state optimal choice at any history.

Finally, the integrand in Equation~\ref{eq:decision-refinement} is a bounded nonnegative random variable. Its expectation is zero exactly when it is zero almost surely, and positive exactly when it is positive on an event of positive probability. Positivity is precisely the event that $m_q^*(S,C)$ fails to minimize the refined conditional losses, proving the strictness statement.
\end{proof}

The equality condition allows tied optimal methods. More explicitly, equality holds if and only if almost every coarse state--context pair admits a method that remains optimal with conditional probability one over its refined subclasses. In terms of conditional probabilities, this condition is
\[
 \max_{m\in\mathcal M}
 \Pr\!\left(
 r'_m(S',C)=\min_{k\in\mathcal M}r'_k(S',C)
 \,\middle|\, S,C\right)=1
 \quad\text{almost surely}.
\]
To see necessity, equality in Equation~\ref{eq:decision-refinement} makes the coarse optimal choice a refined minimizer almost surely. Conditioning that event on $(S,C)$ gives a method with conditional probability one of remaining optimal. For sufficiency, choose the first method with that conditional probability equal to one. The finite dictionary makes this choice measurable. It is a coarse selector attaining the refined optimal risk, which forces equality by the proposition. Consequently, equality does not require identical refined loss vectors. It requires a common optimal choice within each coarse information class. If equality holds, every coarse optimal selector has zero refined excess loss by the same identity.

Comparison with the full observed history gives a related exact decomposition,
\begin{equation}
 \mathfrak R(q)-\mathfrak R_H
 =\mathbb E\!\left[
 \ell(m_q^*(S,C),H,C)-\min_{m\in\mathcal M}\ell(m,H,C)\right].
 \label{eq:aliasing-regret}
\end{equation}
Thus $\mathfrak R(q)=\mathfrak R_H$ exactly when an optimal state-limited choice is history-optimal almost surely. This characterizes decision sufficiency under the fixed loss and distribution.

\subsubsection{Unresolved procedural ambiguity}

Equation~\ref{eq:aliasing-regret} identifies the loss associated with retaining too little history information. To quantify this loss through disagreement about the preferred method, assume that the history-optimal method is unique and separated from the alternatives by a positive loss margin.

\begin{corollary}[Loss forced by aliased procedural choices]
\label{cor:aliasing-margin}
Suppose the history-optimal method
$J=\operatorname*{arg\,min}_{m\in\mathcal M}\ell(m,H,C)$
is unique almost surely. Assume there is $\gamma>0$ such that, almost surely, every $m\ne J$ satisfies
$\ell(m,H,C)-\ell(J,H,C)\geq\gamma$.
Define
\begin{equation}
 A(q)=\mathbb E\!\left[1-\max_{m\in\mathcal M}\Pr(J=m\mid S,C)\right].
 \label{eq:choice-ambiguity}
\end{equation}
Then
\begin{equation}
 \gamma A(q)\leq\mathfrak R(q)-\mathfrak R_H\leq A(q).
 \label{eq:aliasing-margin}
\end{equation}
\end{corollary}

\begin{proof}
Let $p_m(S,C)=\Pr(J=m\mid S,C)$. For any deterministic state-limited choice $\widehat J$,
\[
 \Pr(\widehat J\ne J\mid S,C)
 =1-p_{\widehat J}(S,C)
 \geq 1-\max_{m\in\mathcal M}p_m(S,C).
\]
The same bound holds for a randomized choice with probabilities $a(m\mid S,C)$ because its conditional probability of matching $J$ is
$\sum_m a(m\mid S,C)p_m(S,C)\leq\max_m p_m(S,C)$.
A deterministic choice $\widetilde J$ of a conditionally most probable method attains equality. Ties are resolved by the fixed ordering of $\mathcal M$, so $\widetilde J$ is measurable. Taking expectations proves that $A(q)$ is the smallest probability of selecting a method different from $J$ using $(S,C)$.

By the margin assumption, every choice satisfies the pointwise inequality
\[
 \ell(\widehat J,H,C)-\ell(J,H,C)
 \geq\gamma\mathbf 1\{\widehat J\ne J\}.
\]
In particular, applying it to $m_q^*$ and using Equation~\ref{eq:aliasing-regret} gives
\[
 \mathfrak R(q)-\mathfrak R_H
 \geq\gamma\Pr(m_q^*(S,C)\ne J)
 \geq\gamma A(q),
\]
which proves the lower bound.

For the upper bound, use the conditionally most probable method $\widetilde J$. Its excess loss is zero when $\widetilde J=J$, and at most one otherwise because all losses lie in $[0,1]$. Since $m_q^*$ minimizes expected continuation loss among state-limited selectors,
\[
 \begin{aligned}
 \mathfrak R(q)-\mathfrak R_H
 &\leq\mathbb E[\ell(\widetilde J,H,C)-\ell(J,H,C)]\\
 &\leq\Pr(\widetilde J\ne J)
 =A(q).
 \end{aligned}
\]
The selector minimizing misclassification need not minimize continuation loss. Its role in this argument is to provide a feasible state-limited choice with controlled loss.
\end{proof}

Under the margin assumption, $A(q)>0$ forces positive excess risk even for the optimal state-limited selector. Conversely, $A(q)=0$ means that $J$ is determined almost surely by $(S,C)$ and gives zero excess risk. Improving the selector while holding its information fixed cannot remove this ambiguity.

The bounds are sharp. Fix a common context and two histories that $q$ maps to the same state, with probabilities $\alpha$ and $1-\alpha$, where $0<\alpha<1$. For two methods with respective loss vectors $(0,\gamma)$ and $(\gamma,0)$ across the two histories, where $0<\gamma\leq1$, the full-history risk is zero and
\[
 A(q)=\min(\alpha,1-\alpha),\qquad
 \mathfrak R(q)-\mathfrak R_H
 =\gamma\min(\alpha,1-\alpha)=\gamma A(q).
\]
Thus the lower bound is attained for every such margin. Taking $\gamma=1$ attains the upper bound as well, so both constants can be attained under the stated assumptions.

For a preparation example, fix a task and role context in which the required object is held and its preparation status is observed. The coarse state records only possession, with probability $\alpha\in(0,1)$ of an unprepared object. Two methods have normalized continuation losses
\[
 \begin{array}{c|cc}
  &\text{unprepared}&\text{prepared}\\ \hline
  \text{prepare-then-deliver}&\varepsilon&\varepsilon\\
  \text{direct delivery}&1&0
 \end{array}
 \qquad 0<\varepsilon<1.
\]
Preparation incurs a positive execution cost even when unnecessary.

The coarse conditional expected losses are $\varepsilon$ and $\alpha$, respectively. Hence its optimal risk is $\mathfrak R(q)=\min(\varepsilon,\alpha)$. A refinement recording the observed preparation status selects prepare-then-deliver only for unprepared histories and direct delivery otherwise. It attains
\[
 \mathfrak R(q')=\mathfrak R_H=\alpha\varepsilon,
\]
and the improvement is
\[
 \mathfrak R(q)-\mathfrak R(q')=
 \begin{cases}
   \alpha(1-\varepsilon),&\alpha\leq\varepsilon,\\
   \varepsilon(1-\alpha),&\alpha>\varepsilon.
 \end{cases}
\]
The improvement is strictly positive, including at the coarse-choice tie $\alpha=\varepsilon$. The uniquely optimal methods differ across histories, with margin $\min(\varepsilon,1-\varepsilon)$ and unresolved choice probability $\min(\alpha,1-\alpha)$. If preparation always succeeded at zero relevant cost, prepare-then-deliver would be optimal in both histories, so the same physical distinction would have zero decision value.

\subsubsection{Historical reassignment and procedure connections}

We now fix a finite set $L$ of historical transition occurrences whose endpoints support the procedural memory. Each $\ell\in L$ has fixed source and target boundary occurrences $\omega_\ell^-$ and $\omega_\ell^+$ in a boundary set $\Omega$. A boundary occurrence identifies a particular boundary in its original execution, rather than merely the text of an observation. The projections $q:\Omega\to\mathcal X_\Sigma$ and $q':\Omega\to\mathcal X_{\Sigma'}$ assign complete states consistently whenever the same boundary is referenced. In this part of the analysis, $q$ and $q'$ act on the fixed boundary records, and no boundary is excluded.

Define candidate occurrence connections and continuations sharing an identical boundary by
\begin{equation}
 \begin{aligned}
 J_q&=\{(\ell,k)\in L^2:q(\omega_\ell^+)=q(\omega_k^-)\},\\
 W&=\{(\ell,k)\in L^2:\omega_\ell^+=\omega_k^-\}.
 \end{aligned}
 \label{eq:joins}
\end{equation}
The ordered pairs in $J_q$ share an intermediate abstract state and may come from different task instances. Those in $W$ share the same recorded boundary.

Assume that $q=\pi\circ q'$ on $\Omega$ for a map $\pi:q'(\Omega)\to q(\Omega)$. For each coarse state $s$, let $\mathcal J_s=\{j\in q'(\Omega):\pi(j)=s\}$ denote its refined subclasses. Only subclasses used by an endpoint contribute to the following finite counts,
\[
 a_{sj}=\bigl|\{\ell\in L:q'(\omega_\ell^+)=j\}\bigr|,
 \qquad
 b_{sj}=\bigl|\{\ell\in L:q'(\omega_\ell^-)=j\}\bigr|,
 \quad j\in\mathcal J_s.
\]
Thus $a_{sj}$ counts incoming occurrences and $b_{sj}$ counts outgoing occurrences at refined state $j$ within coarse state $s$. All sums below can be restricted to endpoint assignments appearing in the finite pool.

\begin{proposition}[Refining historical state assignments]
\label{prop:interfaces}
Under the fixed-occurrence and refinement assumptions above,
$W\subseteq J_{q'}\subseteq J_q$.
The number of candidate connections removed at an old state $s$ is
\begin{equation}
 \begin{aligned}
 \Delta_s
 &=\Bigl(\sum_{j\in\mathcal J_s}a_{sj}\Bigr)
   \Bigl(\sum_{j\in\mathcal J_s}b_{sj}\Bigr)
   -\sum_{j\in\mathcal J_s}a_{sj}b_{sj}\\
 &=\sum_{\substack{i,j\in\mathcal J_s\\i\ne j}}a_{si}b_{sj}.
 \end{aligned}
 \label{eq:join-count}
\end{equation}
Moreover, $|J_q\setminus J_{q'}|=\sum_s\Delta_s$.
\end{proposition}

\begin{proof}
If $(\ell,k)\in J_{q'}$, then $q'(\omega_\ell^+)=q'(\omega_k^-)$. Applying $\pi$ to the two equal assignments gives $q(\omega_\ell^+)=q(\omega_k^-)$, hence $(\ell,k)\in J_q$. This proves $J_{q'}\subseteq J_q$. If $(\ell,k)\in W$, both endpoints name the same boundary occurrence. Consistency of $q'$ assigns them the same refined state, so $(\ell,k)\in J_{q'}$. This proves $W\subseteq J_{q'}$.

At coarse state $s$, every occurrence ending at $s$ can be paired with every occurrence starting at $s$. There are $\sum_j a_{sj}$ incoming occurrences and $\sum_j b_{sj}$ outgoing occurrences, so their Cartesian product has size $(\sum_j a_{sj})(\sum_j b_{sj})$. Under $q'$, equality of the intermediate states permits only pairs within the same subclass. The product for subclass $j$ has size $a_{sj}b_{sj}$, and these products are disjoint. Subtracting their total size gives the first expression for $\Delta_s$. Expanding the product and removing its diagonal terms gives the second expression.

Each removed pair has a unique coarse intermediate state. The removed pairs are therefore partitioned by $s$, and summing their counts gives $|J_q\setminus J_{q'}|=\sum_s\Delta_s$.
\end{proof}

For any fixed sequence $(\ell_1,\ldots,\ell_n)$ with $n\geq2$ and
\[
 \omega_{\ell_i}^+=\omega_{\ell_{i+1}}^-
 \quad\text{for every }i=1,\ldots,n-1,
\]
every adjacent pair lies in $W$ and hence in $J_{q'}$. Such a fixed occurrence chain remains connected after refinement. Conversely, every chain connected by $J_{q'}$ is also connected by $J_q$, by applying the other inclusion to each adjacent pair. This result characterizes connectivity within the fixed historical occurrence pool.

Equation~\ref{eq:join-count} counts cross-subclass pairs of historical occurrences whose shared abstract endpoint is separated by refinement.

Historical reassignment also determines where a previously learned method can retain its evidence. For an old method-bearing transition $e$, let $\operatorname{supp}(e)$ be its finite set of supporting realizations. After a schema revision, let $\psi(\omega)$ be the retained state identity assigned to boundary $\omega$, or $\bot$ if the occurrence is excluded. This reassignment need not be a refinement. Recall the complete projected support set from Equation~\ref{eq:method-migration},
\[
 \Pi_b(e)=\{(\psi(\omega_\ell^-),\psi(\omega_\ell^+)):
                   \ell\in\operatorname{supp}(e)\}.
\]
The method can retain its entire nonempty support association at a single valid state pair if and only if
\[
 \Pi_b(e)=\{(u,v)\},\qquad u,v\ne\bot,\quad u\ne v.
\]
For necessity, retaining the entire nonempty association at one pair means that every supporting realization has the same projected source $u$ and target $v$. The complete set of projected pairs must therefore be the displayed singleton. The pair must have nonexcluded, distinct endpoints because these are required for a retained transition. For sufficiency, a singleton with valid endpoints assigns every supporting realization to the common pair, so attaching the method to that pair preserves the entire support association. Empty support, any excluded endpoint, or a coincident-endpoint singleton cannot satisfy this criterion. In particular, removing invalid pairs before forming the set would no longer test preservation of the entire support.

If support separates into multiple pairs, direct migration cannot preserve the entire association at one pair. The learner must reassess the evidence at its supported pairs. Support at $(u_0,v_0)$ and $(u_1,v_1)$, with all four endpoints distinct, supplies no association at the cross-pairs $(u_0,v_1)$ and $(u_1,v_0)$. The criterion specifies where a method's historical evidence is retained after reassignment.

\end{document}